\documentclass[runningheads]{llncs}
\usepackage{graphicx}
\usepackage{tikz}
\usepackage{comment}
\usepackage{amsmath,amssymb} % define this before the line numbering.
\usepackage{color}
\usepackage{orcidlink}
\usepackage{bm}
\usepackage{subcaption}
\usepackage{colortbl}
\usepackage{algorithm}
\usepackage{algpseudocode}
\usepackage{multirow}
\usepackage{wrapfig,lipsum,booktabs}
\usepackage[misc]{ifsym}
\usepackage[accsupp]{axessibility}  % Improves PDF readability for those with disabilities.
\usepackage{lipsum}

\newcommand\blfootnote[1]{%
  \begingroup
  \renewcommand\thefootnote{}\footnote{#1}%
  \addtocounter{footnote}{-1}%
  \endgroup
}

\begin{document}
% \linenumbers
\pagestyle{headings}
\mainmatter
\def\ECCVSubNumber{4295}  % Insert your submission number here

\title{Learning to Weight Samples for Dynamic Early-exiting Networks} % Replace with your title

\begin{comment}
\titlerunning{ECCV-22 submission ID \ECCVSubNumber} 
\authorrunning{ECCV-22 submission ID \ECCVSubNumber} 
\author{Anonymous ECCV submission}
\institute{Paper ID \ECCVSubNumber}
\end{comment}
%******************

%\begin{comment}
\titlerunning{L2W-DEN}

\author{Yizeng Han\inst{1}$^*$\orcidlink{0000-0001-5706-8784} \and
Yifan Pu\inst{1}$^*$\orcidlink{0000-0002-0404-1737} \and
Zihang Lai\inst{2}$^\dag$\orcidlink{0000-0002-9872-0756} \and
Chaofei Wang\inst{1}\orcidlink{0000-0002-3678-691X} \and
Shiji Song\inst{1}\orcidlink{0000-0001-7361-9283} \and
Junfeng Cao\inst{3}\orcidlink{0000-0002-2101-7974} \and
Wenhui Huang\inst{3}\orcidlink{0000-0001-8123-2441} \and
Chao Deng\inst{3}\orcidlink{0000-0003-4449-5247} \and
Gao Huang\inst{1}\textsuperscript{(\Letter)}\orcidlink{0000-0002-7251-0988}
}

\authorrunning{Y. Han, Y. Pu et al.}
\institute{Tsinghua University, Beijing 100084, China\\
\email{\{hanyz18, pyf20, wangcf18\}@mails.tsinghua.edu.cn}\blfootnote{$*$ Equal contribution. \Letter~Corresponding author.}\\
\email{\{shijis, gaohuang\}@tsinghua.edu.cn}\and
Carnegie Mellon University, Pennsylvania 15213, United States
\email{zihangl@andrew.cmu.edu}\blfootnote{$\dag$~Work done during an internship at Tsinghua University.}\and
China Mobile Research Institute, Beijing 100084, China\\
\email{\{caojunfeng, huangwenhui, dengchao\}@chinamobile.com}}
%\end{comment}
%******************

\maketitle
%!TEX root = _main.tex
\begin{abstract}
   Early exiting is an effective paradigm for improving the inference efficiency of deep networks. By constructing classifiers with varying resource demands (the exits), such networks allow \emph{easy} samples to be output at early exits, removing the need for executing deeper layers. While existing works mainly focus on the architectural design of multi-exit networks, the training strategies for such models are largely left unexplored. The current state-of-the-art models treat all samples \emph{the same} during training. However, the \emph{early-exiting} behavior during testing has been ignored, leading to a gap between training and testing. In this paper, we propose to bridge this gap by \emph{sample weighting}. Intuitively, \emph{easy} samples, which generally exit early in the network during inference, should contribute more to training early classifiers. The training of \emph{hard} samples (mostly exit from deeper layers), however, should be emphasized by the late classifiers. Our work proposes to adopt a \emph{weight prediction network} to weight the loss of different training samples at each exit. This weight prediction network and the backbone model are jointly optimized under a \emph{meta-learning} framework with a novel optimization objective. By bringing the adaptive behavior during inference into the training phase, we show that the proposed weighting mechanism consistently improves the trade-off between classification accuracy and inference efficiency. Code is available at \url{https://github.com/LeapLabTHU/L2W-DEN}.
   
\end{abstract}
%!TEX root = _main.tex
% \vskip -1in
\section{Introduction}\label{sec:intro}
% \vskip -0.05in
Although significant improvements have been achieved by deep neural networks in computer vision \cite{alexnet,Simonyan15,szegedy2015going,he2016deep,huang2017densely,dosovitskiy2020image,liu2021swin}, the high computational cost of deep models still prevents them from being applied on resource-constrained platforms, such as mobile phones and wearable devices. Improving the inference efficiency of deep learning has become a research trend. Popular solutions include lightweight architecture design \cite{howard2017mobilenets,zhang2018shufflenet}, network pruning \cite{lecun1990optimal,li2016pruning,liu2017learning,yang2021condensenet}, weight quantization \cite{hubara2016binarized,han2015deepcompression}, and dynamic neural networks \cite{han2021dynamic,panda2016conditional,huang2017multi,yang_resolution_2020,lin2017runtime,bejnordi2019batch,wang2018skipnet,verelst_dynamic_2020,SAR_TIP}.

\begin{figure}[t]
    \centering
    \includegraphics[width=0.85\linewidth]{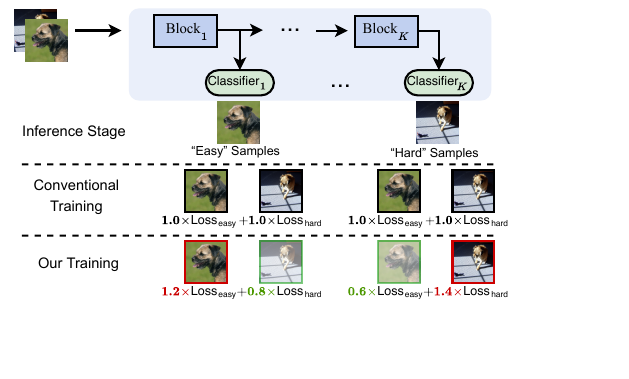}
    % \vskip -0.1in
    \caption{\textbf{Our sample weighting strategy}. At test time, the classifiers at varying depths handle inputs with different complexity (\emph{top}). However, the conventional training strategy (\emph{middle}) treats all the samples equally at multiple exits. In contrast, our weighting mechanism (\emph{bottom}) enables each classifier to emphasize the samples that it is responsible for.}
    \label{fig1}
    % \vskip -0.3in
\end{figure}

Dynamic networks have attracted considerable research interests due to their favorable efficiency and representation power \cite{han2021dynamic}. In particular, they perform a data-dependent inference procedure, and different network components (e.g. layers \cite{wang2018skipnet} or channels \cite{bejnordi2019batch}) could be conditionally skipped based on the input complexity. A typical adaptive inference approach is \emph{early exiting} \cite{huang2017multi,yang_resolution_2020}, which can be achieved by constructing a deep network with multiple intermediate classifiers (early exits). Once the prediction from an early exit satisfies a certain criterion (e.g. the classification \emph{confidence} exceeds some threshold), the forward propagation is terminated, and the computation of deeper layers will be skipped. 
% Representatively, multi-scale dense network (MSDNet) \cite{huang2017multi} performs dynamic early exiting based on the classification confidence of early predictions. 
Compared to the conventional static models, such an \emph{adaptive} inference mechanism can significantly improve the computational efficiency without sacrificing accuracy. When making predictions with shallow classifiers for canonical (\emph{easy}) samples, a substantial amount of computation will be saved by skipping the calculation of deep layers. Moreover, the network is capable of handling the non-canonical (\emph{hard}) inputs with deep exits (Fig.~\ref{fig1}, \emph{Inference Stage}).

Existing works on dynamic networks mainly focus on designing more advanced multi-exit architectures \cite{huang2017multi,yang_resolution_2020}. A naive training strategy is generally adopted by summing up the cross-entropy losses obtained from all classifiers. More importantly, the loss of both \emph{easy} and \emph{hard} samples contributes equally to the final optimization objective (Fig.~\ref{fig1}, \textit{Conventional Training}), regardless of where a sample may \textit{actually exit}. However, these exits at varying depths have different capabilities, and they are responsible for recognizing samples of varying complexity at test time. Such an adaptive inference behavior has been neglected by the naive training paradigm adopted in previous works \cite{huang2017multi,li2019improved,yang_resolution_2020}.

% of dynamic models, and is suboptimal to the trade-off between network performance and inference efficiency.

In this paper, we propose to bridge the gap between training and testing by 
% consider the training of multi-exit models from a novel perspective. In contrast to the existing works that encourage each exit to classify samples of all difficulties, we 
imposing sample-wise weights on the loss of multiple exits. Our motivation is that every classifier is required to handle a \emph{subset} of samples in the adaptive inference scenario. Specifically, the early classifiers only need to recognize some canonical inputs, while the deep classifiers are usually responsible for those non-canonical samples. Therefore, an ideal optimization objective should encourage each exit to emphasize different training samples by weighting their loss. 
% From the optimization perspective, feeding easy samples to low-capacity networks (i.e. classifiers attached at shallow layers) may smooth the learning objective, therefore benefiting their generalization ability \cite{wang2021survey_cl}. 
In a nutshell, the challenge for sample weighting in multi-exit models is two-fold: 1) the exiting decisions are made during \emph{inference}, and we have no prior knowledge of where a specific sample exits; 2) setting proper weights is a non-trivial problem. 
% Due to that , a challenge for optimizing dynamic networks is that the target subset for every exit cannot be obtained in advance.  

To address these challenges, we propose to automatically \emph{learn} appropriate weights by leveraging a \emph{weight prediction network} (WPN). The WPN takes the training loss from all exits as input, producing the weights imposed on the samples at every exit. We jointly optimize the backbone model and the WPN in a \emph{meta-learning} manner. A novel optimization objective is constructed to guide the meta-learning procedure. Precisely, we mimic the test-time early-exiting process to find where the samples will exit during inference. The meta objective for each classifier is then defined as the loss only on the samples that \textit{actually} exit at this classifier. Compared to the conventional training strategy, our specialized meta-objective encourages every exit to impose proper weights on different samples for improved performance in dynamic early exiting (Fig.~\ref{fig1}, \textit{Our Training}).

We evaluate our method on image recognition tasks in two settings: a class balanced setting and a class imbalanced setting. The experiment results on CIFAR\cite{Krizhevsky09learningmultiple}, ImageNet \cite{deng2009imagenet}, and the long-tailed CIFAR \cite{cui2019class} demonstrate that the proposed approach consistently improves the trade-off between accuracy and inference efficiency for state-of-the-art early-exiting networks.
\section{Related works}
\label{sec:related_work}
% \vskip -0.05in
\noindent\textbf{Dynamic early-exiting networks.}
Early exiting is an effective dynamic inference paradigm, allowing \emph{easy} samples to be output at intermediate classifiers. Existing works mainly focus on designing more advanced architectures. For instance, BranchyNet \cite{teerapittayanon2016branchynet} attaches classifier heads at varying depths in an AlexNet \cite{alexnet}. An alternative option is cascading multiple CNNs (i.e. AlexNet \cite{alexnet}, GoogleNet \cite{szegedy2015going} and ResNet \cite{he2016deep}) to perform early exiting \cite{bolukbasi2017adaptive}. It is observed \cite{huang2017multi} that the classifiers may interfere with each other in these chain-structured or cascaded models. This issue is alleviated via dense connection and multi-scale structure \cite{huang2017multi}. Resolution adaptive network (RANet) \cite{yang_resolution_2020} further performs early exiting by conditionally activating high-resolution feature representations.

% The aforementioned works have made extensive efforts on designing multi-exit architectures. However, they typically follow the training routines developed for static networks.
While considerable efforts have been made on the architecture design, the \emph{training} of these models still follows the routines developed for static networks. The \emph{multi-exit} structural property and the \emph{early-exiting} behavior are usually ignored.
Some training techniques are studied to boost the knowledge transfer among exits, yet still neglecting the \emph{adaptive inference} paradigm \cite{li2019improved}. In this paper, we put forth a novel optimization \emph{objective}, encouraging each exit to focus on the samples that they would probably handle in dynamic inference.

\noindent\textbf{Sample weighting.} Different training samples have unequal importance. The idea of sample weighting could be traced back to dataset resampling \cite{chawla2002smote} or instance reweighting \cite{zadrozny2004learning}. These traditional approaches evaluate the sample importance with some prior knowledge. Recent works manage to establish a loss-weight mapping function \cite{freund1997decision,malisiewicz2011ensemble,lin2017focal,johnson2019survey,kumar2010self,jiang2014easy}. There are mainly two directions: one tends to impose larger weights on \emph{hard} samples to deal with the {data imbalance} problem (e.g. focal loss \cite{lin2017focal}); the other focuses more on \emph{easy} samples to reduce the impact brought by {noisy labels} (e.g. self-paced learning \cite{kumar2010self,jiang2014easy}). 

We study a different case: the test-time data distributions for different exits are divergent. Such distributions \emph{could not be obtained in advance} due to the data-dependent inference procedure. The increased number of classifiers further raises challenges for designing the sample-weight mapping function. With our specially designed optimization objective and the meta-learning algorithm, we effectively produce proper sample weights for training the multi-exit model.

\noindent\textbf{Meta-learning in sample weighting.} Due to its remarkable progress, meta-learning \cite{hospedales2020meta,finn2017model} has been extensively studied in sample weighting \cite{l2rw,meta_weight_net,zhang2021learning}. Existing approaches mainly focus on tackling class imbalance or corrupted label problems. In contrast, our goal is to improve the \emph{inference efficiency} of early-exiting networks in a more general setting, without any assumption on the dataset.

\section{Method}\label{sec:method}
% \vskip -0.1in
In this section, we first introduce the preliminaries of dynamic early-exiting networks and their conventional training strategies. Then the sample weighting mechanism and our meta-learning algorithm will be presented. 
% \vskip -0.2in
\subsection{Preliminaries}
% \vskip -0.05in
\noindent\textbf{Multi-exit networks.} A typical approach to setting up a $K$-exit network is attaching $K\!-\!1$ intermediate classifiers at varying depths of a deep model \cite{huang2017multi,yang_resolution_2020}. For an input sample $\mathbf{x}$, the prediction from the $k$-th exit can be written as
\begin{align}
    \hat{y}^{(k)}=\arg\max_{c} {p}^{(k)}_c=\arg\max_{c} f^{(k)}_c(\mathbf{x}; \mathbf{\mathbf{\Theta}}_f^{(k)}), k=1,2,\cdots, K,
\end{align}
\noindent where $p^{(k)}_c$ is the $k$-th classifier's output probability for class $c$, and $f^{(k)}(\cdot; {\mathbf{\Theta}}_f^{(k)})$ represents the $k$-th sub-network with parameter ${\mathbf{\Theta}}_f^{(k)}$. Note that the parameters of different sub-networks are shared in a nested way except the classifiers. We denote the whole classification model as $f$ and its parameters as $\mathbf{\Theta}_f$.

\noindent\textbf{Dynamic early exiting.} Extensive efforts have been made to perform early exiting based on multi-exit models. A typical approach is terminating the forward propagation \emph{once} the classification \emph{confidence} ($\max_{c} {p}^{(k)}_c$) at a certain exit exceeds a given threshold ($\epsilon_k$). The final prediction is obtained by
\begin{align}\label{early_exit_inference_procedure}
    \hat{y}=\hat{y}^{(k)}, \text{ if }\max_{c} {p}^{(k)}_c \ge \epsilon_k, \text{ and }\max_{c} {p}^{(j)}_c < \epsilon_j, \forall j\le k, k\le K-1.
\end{align}
The predictions $\hat{y}^{(k)} (k\! =1,2,\cdots, K)$ are obtained sequentially before satisfying the criterion in Eq.~(\ref{early_exit_inference_procedure}) or reaching the last exit. The threshold for exit-$k$ ($\epsilon_k$) can be decided on a validation set according to the computational budget.

\noindent\textbf{Conventional training methods.} A naive training strategy adopted by existing works \cite{huang2017multi,yang_resolution_2020} is directly minimizing a cumulative loss function:
% \vskip -0.2in
\begin{align}\label{baseline_train}
    \mathcal{L} &=\sum\nolimits_{k=1}^{K} \frac{1}{N}\sum\nolimits_{i=1}^N l^{(k)}_i \triangleq\sum\nolimits_{k=1}^{K} \frac{1}{N}\sum\nolimits_{i=1}^N\mathrm{CE}(f^{(k)}(\mathbf{x}_i; \mathbf{\mathbf{\Theta}}_f^{(k)}), y_i), %\nonumber \\
    % l^{(k)}_i&\triangleq\mathrm{CE}(f^{(k)}(\mathbf{x}_i; \mathbf{\mathbf{\Theta}}_f^{(k)}), y_i),
    % \nonumber\\ &=\sum\nolimits_{k=1}^{K}\lambda^{(k)}\frac{1}{N}\sum\nolimits_{i=1}^N\mathrm{CE}(f^{(k)}(\mathbf{x}_i; \mathbf{\mathbf{\Theta}}_f^{(k)}), y_i),
\end{align}
\noindent where CE is the cross-entropy loss, and $N$ is the number of training samples. 
% $l^{(k)}_i\!=\!\mathrm{CE}(f^{(k)}(\mathbf{x}_i; \mathbf{\mathbf{\Theta}}_f^{(k)}), y_i)$ is the cross-entropy loss of $(\mathbf{x}_i, y_i)$ at exit $k$, and $N$ denotes the number of training samples. 
% Such a training strategy regarding the multi-exit model as a static network is adopted in most existing works \cite{huang2017multi,yang_resolution_2020,wang2021images} .
% \cite{huang2017multi,yang_resolution_2020,wang2021images} regard the multi-exit model as a static network in training, and simply set $\lambda^{(k)}\! =\!1, k\! =\!1,2,\cdots, K$. 
% The objective in Eq. \ref{baseline_train} encourages all the exits to correctly classify every input.
% different training samples contribute equally to multiple sub-networks, and these sub-networks . 

\subsection{Sample-weighting for early-exiting networks}\label{sec:sample-weighting}

\begin{figure}[t]
    \begin{center}
      \includegraphics[width=\linewidth]{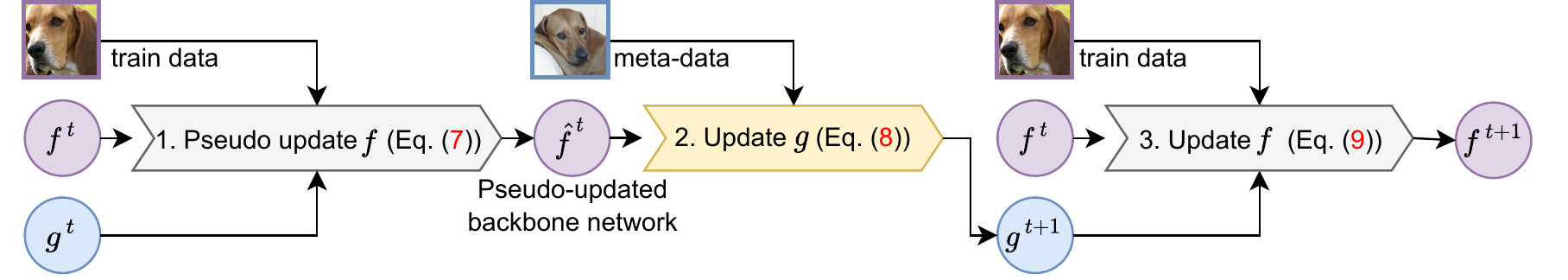}
    \end{center}
    % \vskip -0.25in
    \caption{\textbf{Our training pipeline in iteration $t$}. It consists of 3 steps: 1) the backbone network $f^t$ is \emph{pseudo-updated} to $\hat{f}^t$ using training samples; 2) a meta objective is computed using $\hat{f}^t$ on meta samples. This objective guides the update of the WPN $g^t$; 3) $f^t$ is updated using the new weights predicted by $g^{t+1}$.}
    % \vskip -0.2in
    \label{fig_pipeline}
\end{figure}

In this subsection, we first formulate our sample weighting mechanism, and then introduce the proposed meta-learning objective. The optimization method is further presented. See Fig.~\ref{fig_pipeline} for an overview of the training pipeline.

\noindent\textbf{Sample weighting with weight prediction network.}\label{sec_sample_weighting}
We can observe from Eq.~(\ref{early_exit_inference_procedure}) that test samples are adaptively allocated to multiple exits according to their prediction confidence  during inference. Sub-networks with varying depths are responsible for handling different subsets of samples. Therefore, it is suboptimal to set the same optimization objective for these exits as in Eq.~(\ref{baseline_train}). To this end, we propose to ameliorate the training objective by sample weighting:
\begin{align}\label{weighting_train}
    \mathcal{L} &=\sum\nolimits_{k=1}^{K} \frac{1}{N}\sum\nolimits_{i=1}^N \textcolor{blue}{w^{(k)}_i} l^{(k)}_i.
    % \nonumber\\ &=\sum\nolimits_{k=1}^{K}\lambda^{(k)}\frac{1}{N}\sum\nolimits_{i=1}^N\mathrm{CE}(f^{(k)}(\mathbf{x}_i; \mathbf{\mathbf{\Theta}}_f^{(k)}), y_i),
\end{align}
\noindent\emph{Weight prediction network.} Since we have no prior knowledge of where a specific sample exits, and the function mapping from input to weight is hard to establish, we propose to automatically \emph{learn} the weight $w_i^{(k)}$ from the input $\mathbf{x}_{i}$ by a {weight prediction network} (WPN, denoted by $g$): $\mathbf{w}_i\! =\![w_i^{(1)},w_i^{(2)},\cdots,w_i^{(K)}]\! =\!g(\bm{l}_i;\mathbf{\Theta}_g)$, where $\bm{l}_i\! =\![l_i^{(1)},l_i^{(2)},\cdots,l_i^{(K)}]$ is the training loss for sample $\mathbf{x}_i$ at $K$ exits. The WPN $g$ is established as an MLP with one hidden layer, and $\mathbf{\Theta}_g$ is the parameters of the WPN. We learn the backbone parameters $\mathbf{\Theta}_f$ and the WPN parameters $\mathbf{\Theta}_g$ in a meta-learning manner. Note that the WPN is only used for training, and no extra computation will be brought during the inference stage. 

\noindent\textbf{The meta-learning objective.}\label{sec_meta_objective} We construct a novel optimization objective to bring the test-time adaptive behavior into training.
% \vskip -0.5i

% Then the weight is imposed on the loss $l^{(k)}_i$ for training the multi-exit network. 

% \emph{Weight prediction network.} We propose to build a weight prediction network (WPN) $g(\bm{l}_i;\mathbf{\Theta}_g)$ for mapping an input sample to its weight. The WPN is an MLP with 1 hidden layer. It directly produces the weights $\mathbf{w}_i\!=\![w_i^{(1)},w_i^{(2)},\cdots,w_i^{(K)}]$ based on the training loss from $K$ exits $\bm{l}_i\!=\![l_i^{(1)},l_i^{(2)},\cdots,l_i^{(K)}]$. 

\begin{figure}[t]
    % \vskip -0.1in
      \centering
        % \vskip -0.05in
        \includegraphics[width=\linewidth]{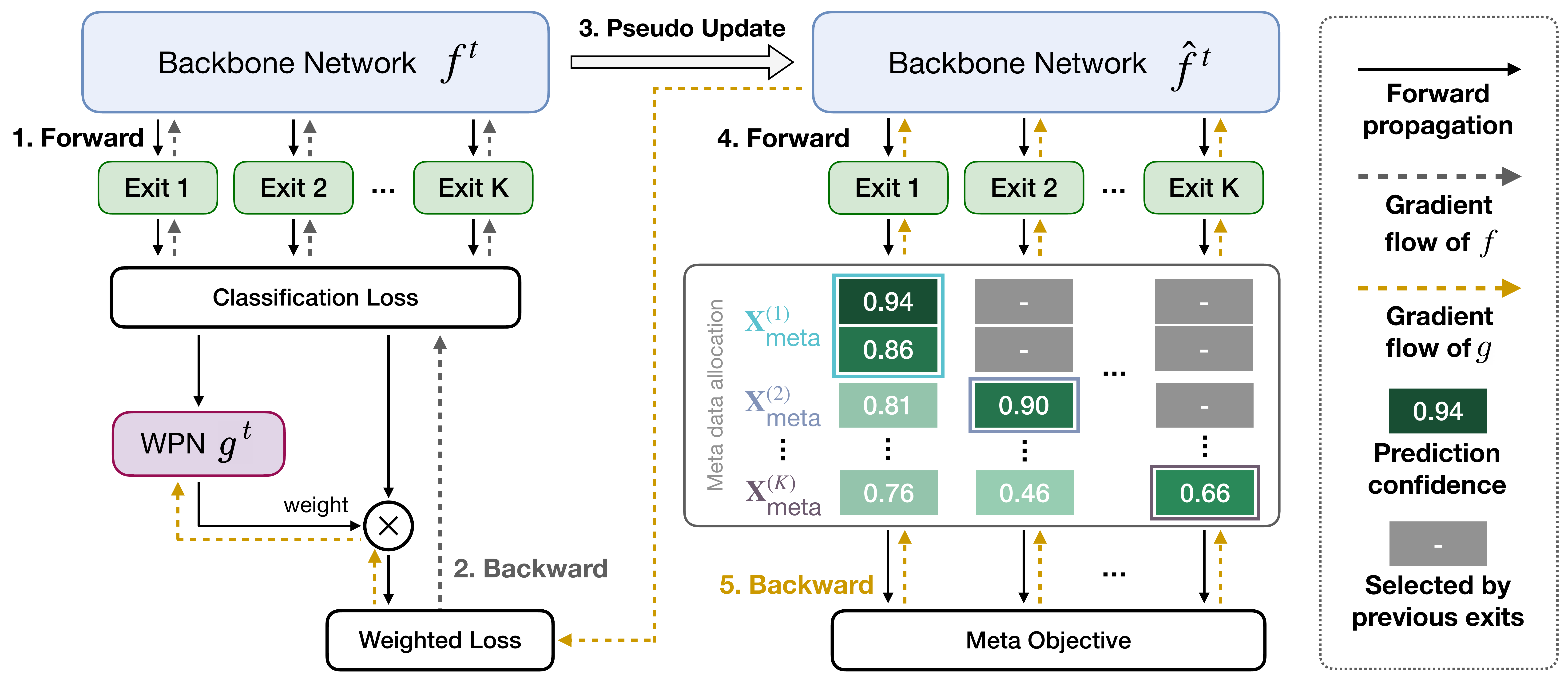}
        % \vskip -0.1in
        \caption{\textbf{Updates of the backbone model $f$ and the WPN $g$ in detail.} First, we compute the weighted classification loss (Eq.~(\ref{optim_net_eq})), which guides the pseudo update on the backbone network $f^t$ (Steps 1, 2, and 3). The meta objective (Eq.~(\ref{meta_loss})) is then computed to update the WPN $g^t$ (Steps 4 and 5).} 
        % The meta data allocation process is shown on the right.}
        %  (see Sec.~\ref{sec_meta_objective}, \emph{The meta-learning objective}).}
        % \vskip -0.25in
        \label{fig_gradient_flow}
\end{figure}

\noindent\emph{Weighted classification loss.} With our weighting scheme (given $\mathbf{\Theta}_g$), the optimization objective of the backbone model parameter $\mathbf{\Theta}_f$ can be written as 
% \begin{align}\label{optim_net_eq}
%      &\mathbf{\Theta}_f^{\star}(\mathbf{\Theta}_g)=\arg\min_{\mathbf{\Theta}_f} \mathcal{L}_{\mathrm{tr}}(\mathbf{\Theta}_f, \mathbf{\Theta}_g) \nonumber\\
%     &\triangleq \arg\min_{\mathbf{\Theta}_f}  \sum\nolimits_{k=1}^K\!\!\frac{1}{N}\!\!\!\sum_{i:\mathbf{x}_{i}\in\mathbf{X}_{\mathrm{tr}}}\!\!\!\!w_i^{(k)}(\mathbf{x}_{i}) l^{(k)}(\mathbf{x}_{i}, y_{i};\mathbf{\Theta}_f),
% \end{align}
% \vskip -0.25in
\begin{align}\label{optim_net_eq}
    \mathbf{\Theta}_f^{\star}(\mathbf{\Theta}_g)&=\arg\min_{\mathbf{\Theta}_f} \mathcal{L}_{\mathrm{tr}}(\mathbf{\Theta}_f, \mathbf{\Theta}_g) \nonumber\\
   &\triangleq \arg\min_{\mathbf{\Theta}_f} \sum_{k=1}^K\frac{1}{N}\sum_{i:\mathbf{x}_{i}\in\mathbf{X}_{\mathrm{tr}}}g^{(k)}(\bm{l}_i;\mathbf{\Theta}_g) \cdot l_i^{(k)}(\mathbf{\Theta}_f),
\end{align}
% \vskip -0.1in
\noindent where $\mathbf{X}_{\mathrm{tr}}$ is the training set. Following \cite{li2021meta}, we 
% adopt $\tanh$ as the activation function of the last layer in $g$, and 
scale the output of our WPN with $0<\delta<1$ to obtain a perturbation $\mathbf{\tilde{w}}\in [-\delta, \delta]^{B\times K}$, where $B$ is the batch size. We further normalize the perturbation $\mathbf{\tilde{w}}$ to ensure the summation of its elements is 0. The final weight imposed on the loss is produced via $\mathbf{w} = 1+\tilde{\mathbf{w}}$. 

% along the \emph{exit} dimension 
% to ensure that for any training sample, the summation of its weight perturbations from all $K$ exits is equal to 0: $\sum\nolimits_{k=1}^K \tilde{w}_i^{(k)}\! =\!0, \forall i$. 

% We could alternatively normalize the perturbation along the \emph{batch} dimension (i.e. $\sum\nolimits_{i=1}^B \tilde{w}_i^{(k)}\! =\!0, \forall k$), or on the overall $B\!\times\!K$ perturbation matrix (i.e. $\sum\nolimits_{i=1}^B\sum\nolimits_{k=1}^K \tilde{w}_i^{(k)}\!=\!0$). We empirically found that normalizing $\mathbf{\tilde{w}}$ along the \emph{exit} dimension leads to higher performance. The experimental results are presented in Sec.~\ref{sec:ablation}.

\noindent\emph{Meta data allocation.} The goal of our weighting mechanism is improving the model performance in the dynamic inference scenario. To construct the optimization objective for the WPN $g$, we first mimic the early exiting procedure on a meta set $\mathbf{X}_\mathrm{meta}$, and obtain the meta samples exiting at different exits $\mathbf{X}^{(k)}_\mathrm{meta}, k\! =\!1,\!2,\!\cdots\!,\!K$. Specifically, we define a budget controlling variable $q$ to decide the number of samples that exit at each exit: $N_k\! =\!\frac{q^k}{\sum\nolimits_{k=1}^K q^k} \times N_{\mathrm{meta}}, q>0,$ where $N_\mathrm{meta}$ is the sample number in the meta set $\mathbf{X}_\mathrm{meta}$. When $q\! =\!1$, the output numbers for different exits are equal, and $q\!>\!1$ means more samples being output by deeper exits, together with more computation. In \emph{training}, we generally tune the variable $q$ on the validation set and fix it in the meta optimization procedure; in \emph{testing}, we vary $q$ from 0 to 2 to cover a wide range of computational budgets, and plot the curves of accuracy-computation in Sec.~\ref{sec:exp} (e.g. Fig.~\ref{msd_cifar_main_results}).

Given $q$, Fig.~\ref{fig_gradient_flow} (right) illustrates the procedure of obtaining $\mathbf{X}^{(k)}_{\mathrm{meta}}$: each exit selects its most confident $N_k$ samples which have not been output at previous classifiers. Note that for the last exit,  $\mathbf{X}^{(K)}_{\mathrm{meta}}$ contains all the unselected samples.

\noindent\emph{Meta objective for dynamic early exiting.} 
% The weighting mechanism is performed to improve the model performance in the scenario of early exiting. Therefore, a meta objective is required for training the parameters of our weight prediction network $\mathbf{\Theta}_g$.  
Instead of correctly recognizing all meta samples with diverse complexity, the meta objective for a specific exit is the classification loss on the samples that \emph{actually exit} at this exit. The overall meta objective is obtained by aggregating the meta objectives from $K$ exits:
% \vskip -0.2in
\begin{equation}\label{meta_loss}
\begin{split}
    \mathbf{\Theta}_g^{\star}&=\arg\min_{\mathbf{\Theta}_g} \mathcal{L}_{\mathrm{meta}}(\mathbf{\Theta}_f^{\star}(\mathbf{\Theta}_g)) \\
    & \triangleq \arg\min_{\mathbf{\Theta}_g} \sum\nolimits_{k=1}^K \!\frac{1}{N_k} \sum_{\textcolor{blue}{j:\mathbf{x}_{j}\in \mathbf{X}^{(k)}_{\mathrm{meta}}}}l^{(k)}(\mathbf{x}_{j},y_{j};\mathbf{\Theta}_f^{\star}(\mathbf{\Theta}_g)).
\end{split}
\end{equation}
%  In our meta objective (Eq.~(\ref{meta_loss})), $\mathbf{X}^{(k)}_{\mathrm{meta}}$ represents the samples allocated to exit-$k$ in the meta-set based on the criterion in Eq.~(\ref{early_exit_inference_procedure}), and $N_k$ is the amount of samples in $\mathbf{X}^{(k)}_{\mathrm{meta}}$. The distribution of $N_k, k=1,2,\cdots,K$ is determined by a variable $q$:

% \begin{wrapfigure}{r}{0.55\linewidth}
%     \begin{center}
%       \includegraphics[width=\linewidth]{output_distribution.pdf}
%     \end{center}
%     \caption{The distribution of the number of output samples over different exits.}
%     \label{dist_q}
%   \end{wrapfigure}

% \vskip -0.1in
\noindent\textbf{The optimization method.}
\label{sec:meta_method}
We jointly optimize the backbone parameter $\mathbf{\Theta}_f$ (Eq.~(\ref{optim_net_eq})) and the WPN parameter $\mathbf{\Theta}_g$ (Eq.  (\ref{meta_loss})) with an online strategy (see Fig.~\ref{fig_pipeline} for an overview of the optimization pipeline). Existing meta-learning methods for loss weighting typically require a standalone meta-set, which consists of class-balanced data or clean labels \cite{l2rw,meta_weight_net}. In contrast, we simply reuse the training data as our meta data. Precisely, at iteration $t$, we first split the input mini-batch into a training batch $\mathbf{X}_{\mathrm{tr}}$ and a meta batch $\mathbf{X}_{\mathrm{meta}}$.
The training batch $\mathbf{X}_{\mathrm{tr}}$ is used to construct the classification loss (Eq.~(\ref{optim_net_eq})) and optimize the model parameters $\mathbf{\Theta}_f$. Next, the meta batch $\mathbf{X}_{\mathrm{meta}}$ is leveraged to compute the meta objective (Eq.~(\ref{meta_loss})) and train the WPN parameters $\mathbf{\Theta}_g$. 

% Fig.~\ref{fig_pipeline} illustrates the training pipeline in each iteration, and the detailed updating procedure is presented as follows.

\noindent\emph{Pseudo update of the backbone network} is conducted using $\mathbf{X}_{\mathrm{tr}}$:
\vskip -0.1in
\begin{equation}\label{pseudo_update}
\mathbf{\hat{
\Theta}}^{t}_f(\mathbf{\Theta}_g)=\mathbf{\Theta}^{t}_f-\alpha \frac{\partial \mathcal{L}_{\mathrm{tr}}(\mathbf{\Theta}_f,\mathbf{\Theta}_g)}{\partial \mathbf{\Theta}_f
}\Bigg|_{\mathbf{\Theta}^{t}_f},
\end{equation}
% \vskip -0.1in
\noindent where $\alpha$ is the learning rate. Note that $\mathbf{\hat{
\Theta}}^{t}_f$ is a function of $\mathbf{\Theta}_g$, and this pseudo update is performed to construct the computational graph for later optimization of $\mathbf{\Theta}_g$. See Fig.~\ref{fig_gradient_flow} (gray dashed lines) for the gradient flow of this pseudo update.

% \begin{figure}[t]
%     \centering
%     %   \vskip -0.05in
%       \includegraphics[width=\linewidth]{fig3_algo.pdf}
%       \vskip -0.15in
%       \caption{\textbf{Main update steps in an iteration.} The proposed learning algorithm consists of three main steps: (a) pseudo update of the backbone parameter $\mathbf{\Theta}_f$, (b) update of the WPN $\mathbf{\Theta}_g$, and (c) actual update of the backbone model $\mathbf{\Theta}_f$.}
%       \label{fig2_algorithm}
%       \vskip -0.2in
%   \end{figure}
\noindent\emph{Update of the weight prediction network} is performed using our meta objective calculated on $\mathbf{X}_{\mathrm{meta}}$. Concretely, we mimic the early exiting procedure on $\mathbf{X}_{\mathrm{meta}}$, and split it into $K$ subsets without intersection $\mathbf{X}_{\mathrm{meta}}\!=\!\{\mathbf{X}_{\mathrm{meta}}^{(1)}\cup\mathbf{X}_{\mathrm{meta}}^{(2)}\cup \cdots\cup\mathbf{X}_{\mathrm{meta}}^{(K)}\}$, where $\mathbf{X}_{\mathrm{meta}}^{(k)}$ contains the meta samples which should be output at exit-$k$ according to the criterion in Eq. (\ref{early_exit_inference_procedure}). See also Fig.~\ref{fig_gradient_flow} (right) for the procedure.

% Instead of improving the accuracy of each exit on the whole meta-set, our objective enables different exits to focus on their \emph{allocated} samples, with the goal of increasing the network performance for dynamic inference.

% Specifically, an exponential distribution for the output sample amount for multiple exits is first calculated based on an input variable $q$. Then $\mathbf{X}_{\mathrm{meta}}^{(k)}, k=1,2,\cdots, K$ can be obtained by selecting the top confident samples for exit-$k$ sequentially.
% \vskip -0.2in

Receiving the partition of the meta-data based on the pseudo updated backbone network, the parameters of our weight prediction model can be updated:
\begin{equation}\label{update_theta_g}
    \mathbf{\Theta}_g^{t+1}=\mathbf{\Theta}_g^{t}-\beta\frac{\partial \mathcal{L}_{\mathrm{meta}}(\mathbf{\hat{
\Theta}}^{t}_f(\mathbf{\Theta}_g))}{\partial \mathbf{\Theta}_g}\Bigg|_{\mathbf{\Theta}_g^{t}},
\end{equation}

\noindent where $\mathcal{L}_{\mathrm{meta}}(\mathbf{\hat{
\Theta}}^{t}(\mathbf{\Theta}_g))$ is the aggregation of the classification loss from each exit on its \emph{allocated} meta data (Eq.~(\ref{meta_loss})), and $\beta$ is the learning rate. The gradient flow for updating $\mathbf{\Theta}_g$ is illustrated in Fig.~\ref{fig_gradient_flow} (the golden dashed lines).
% By optimizing such a meta loss designed specifically for early exiting, our weight prediction model learns to yield appropriate weights for each training sample at different exits (see also our visualization results in Sec.~\ref{sec:visualization}).
% \vskip -0.2in
\begin{algorithm}
    \caption{The meta-learning Algorithm}\label{algo}
    \begin{algorithmic}
    \Require Training data $\mathcal{D}$, batch size $B$, iteration $T$, interval $I$, budget controller $q$.
    \Ensure Backbone network parameters $\mathbf{\Theta}_f$.
    \For{$t=0$ \textbf{to} $T-1$}
        \State $\{\mathbf{X},\mathbf{y}\} \gets \mathrm{SampleMiniBatch}(\mathcal{D},B)$.
        \State Split $\{\mathbf{X},\mathbf{y}\}$ into $\{\mathbf{X}_{\mathrm{tr}},\mathbf{X}_{\mathrm{meta}},\mathbf{y}_{\mathrm{tr}},\mathbf{y}_{\mathrm{meta}}\}$.
        \If{$t\mod I=0$}
            \State Perform a pseudo update for $\mathbf{\hat{\Theta}}_f^{t}$ by Eq.~(\ref{pseudo_update}).
            \State Perform meta data allocation based on $q$.
            \State Update $\mathbf{\Theta}_g^{t+1}$ by Eq.~(\ref{update_theta_g}).
        \EndIf
        \State Update $\mathbf{\Theta}_f^{t+1}$ by Eq.~(\ref{update_theta_f}).
    \EndFor
    \end{algorithmic}
    % \vskip -0.05in
\end{algorithm}

\noindent\emph{Update of the backbone network} is finally realized based on the updated $\mathbf{\Theta}_g^{t+1}$:
% \vskip -0.1in
\begin{equation}\label{update_theta_f}
    \mathbf{\Theta}_f^{t+1} = \mathbf{\Theta}_f^{t} - \alpha \frac{\partial \mathcal{L}_{\mathrm{tr}}(\mathbf{\Theta}_f,\mathbf{\Theta}_g^{t+1})}{\partial \mathbf{\Theta}_f}\Bigg|_{\mathbf{\Theta}^{t}_f}.
\end{equation}

% \noindent\emph{Analysis of our learning method.} 
We summarize the learning algorithm in Algorithm~\ref{algo}. By mimicking the adaptive inference procedure, our novel objective in Eq.~(\ref{meta_loss}) encourages each exit to correctly classify the samples \emph{which are most probably allocated to it} in the early exiting scenario (blue text in Eq.~(\ref{meta_loss})). For example, early exits at shallow layers may focus more on those \emph{easy} samples (see also our visualization results in Sec.~\ref{sec:exp}, Fig.~\ref{fig_vis}). 
% while deep classifiers should take responsibility for the \emph{hard} samples. 
% The results in Sec.~\ref{sec:exp} further validate the effectiveness of such a weighting mechanism.
% Our visualization results in Fig.~\ref{fig_vis} further validate that the meta-learning algorithm successfully trains the WPN to produce exit-specific weights for each sample based on its training loss. Interestingly, the results in Tab. \ref{anytime_imta_ours} and Tab. \ref{ra_anytime} demonstrate that the accuracy of different exits on the \emph{whole} test set could be improved. 
More empirical analysis is presented in Sec.~\ref{sec:exp}.

\vskip -0.3in
\section{Experiments}
\label{sec:exp}
% We perform extensive experiments on image classification and compare our approach with existing methods. We first introduce the experiment settings in Sec.~\ref{sec_experiment_setting}, and then present the ablation studies in Sec.~\ref{sec:ablation} to validate the design choices made in our weighting mechanism. Next, we empirically evaluate our method on CIFAR \cite{Krizhevsky09learningmultiple} (Sec.~\ref{sec_cifar_results}) and ImageNet \cite{deng2009imagenet} (Sec.~\ref{sec_imagenet}). Finally, we evaluate our training algorithm on the long-tailed CIFAR \cite{cui2019class} in Sec.~\ref{sec_imbalance} in a class imbalance setting. We measure a network's performance in terms of the trade-off between accuracy and the average Mul-Add (multiply-add operation). 

In this section, we first conduct ablation studies to validate the design choices made in our weighting mechanism (Sec.~\ref{sec:ablation}). The main results on CIFAR \cite{Krizhevsky09learningmultiple} and ImageNet \cite{deng2009imagenet} are then presented in Sec.~\ref{sec_cifar_results} and Sec.~\ref{sec_imagenet} respectively. Finally, we evaluate our method on the long-tailed CIFAR \cite{cui2019class} in a class imbalance setting (Sec.~\ref{sec_imbalance}). A network's performance is measured in terms of the trade-off between the accuracy and the Mul-Adds (multiply-add operations). We apply our training algorithm to two representative dynamic early-exiting architectures, i.e. multi-scale dense network (MSDNet) \cite{huang2017multi} and resolution adaptive network (RANet) \cite{yang_resolution_2020}. The experimental setup is provided in the supplementary material.

\begin{figure}[t]
  \centering
  \begin{subfigure}[b]{0.32\textwidth}
    \centering
    \includegraphics[width=\textwidth]{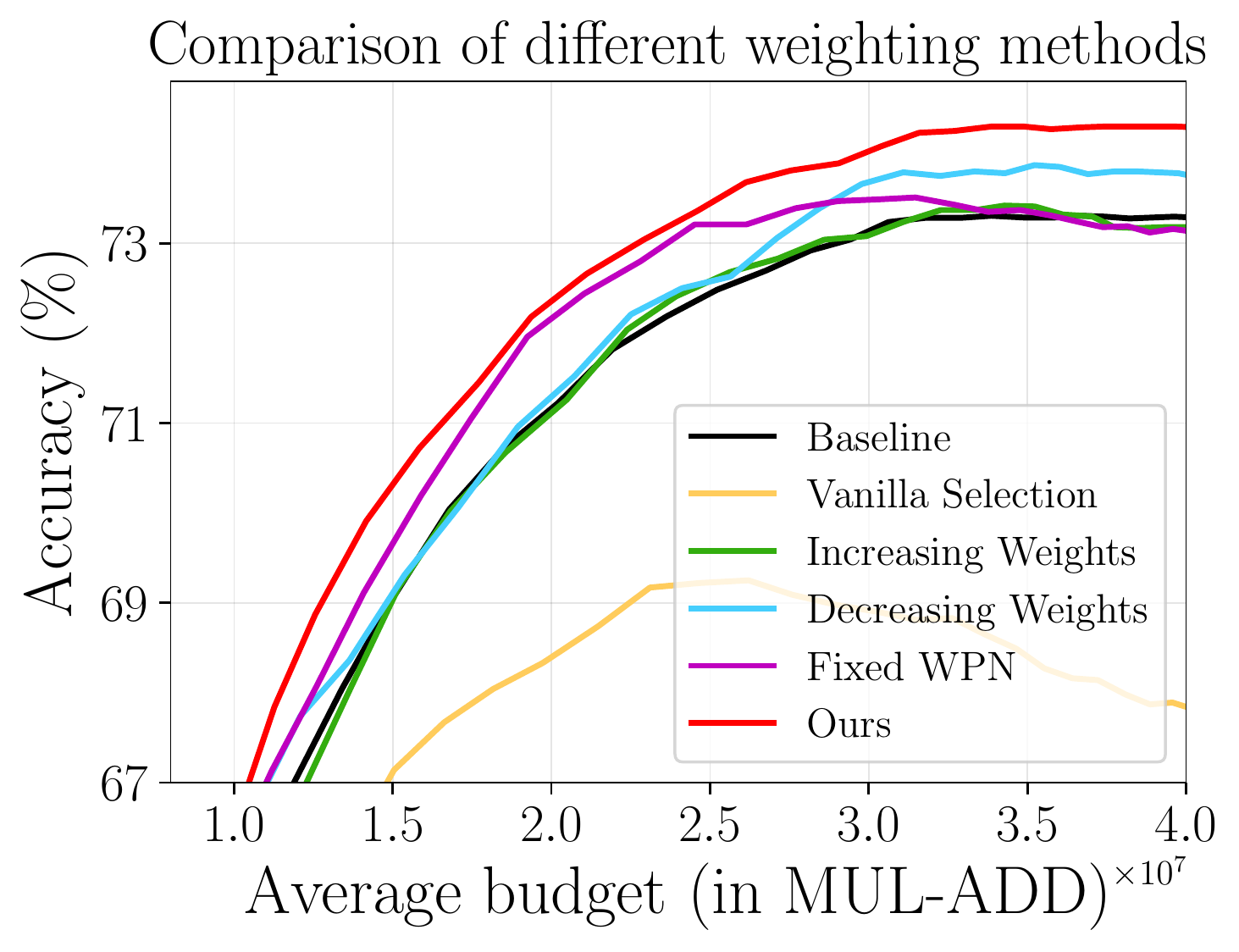}
    % \label{fig:three sin x}
  \end{subfigure}
  % \hfill
  \begin{subfigure}[b]{0.32\textwidth}
      \centering
      \includegraphics[width=\textwidth]{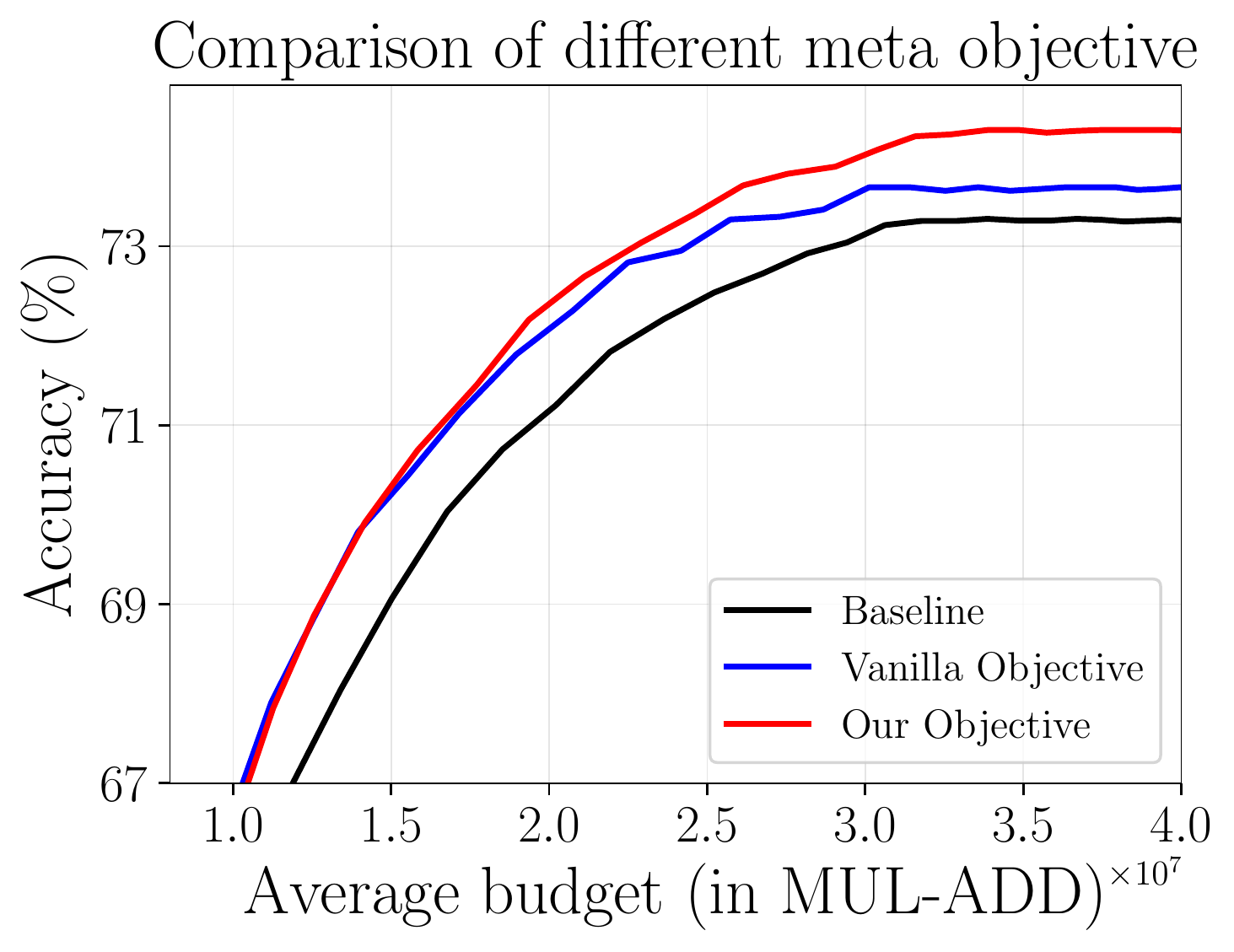}
      % \label{fig:y equals x}
  \end{subfigure}
  % \hfill
  \begin{subfigure}[b]{0.32\textwidth}
      \centering
      \includegraphics[width=\textwidth]{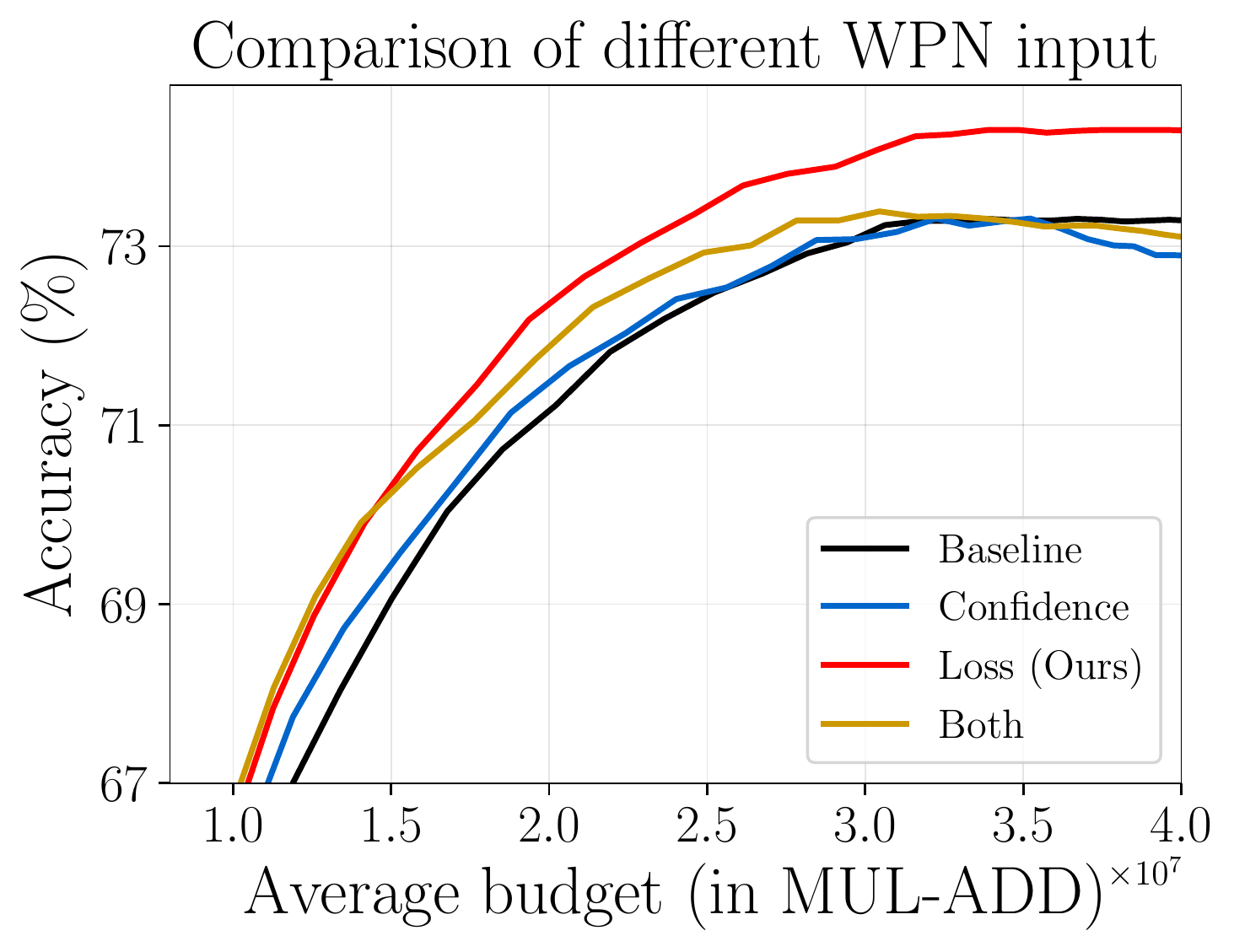}
      % \label{fig:three sin x}
  \end{subfigure}
%   \vskip -0.1in
  \caption{\textbf{Ablation studies on CIFAR-100.} Left: our meta-learning based approach \emph{v.s.} heuristic weighting mechanisms. Middle: the effectiveness of the proposed meta objective. Right: comparison of different WPN inputs.}
  \label{fig:ablation}
%   \vskip -0.25in
\end{figure}

\subsection{Ablation studies}\label{sec:ablation}
% \vskip -0.075in
We perform ablation studies with a 5-exit MSDNet on CIFAR-100 to validate the effectiveness of different settings and variants in our method.

\noindent\textbf{Meta-learning algorithm.} We first verify the \emph{necessity} of our {meta-learning} algorithm by comparing it with three variants: 1) the first replaces the weighting scheme with a vanilla selection scheme: the data allocation (Fig.~\ref{fig_gradient_flow}) is applied directly to the \emph{training} data, and the loss of each exit is only calculated on its \emph{allocated} samples; 2) the second hand-designs a weighting mechanism with \emph{fixed} weights increasing/decreasing with the exit index; 3) the third uses a \emph{frozen} convergent WPN to weight the samples. For the second variant, the weights are set from 0.6 to 1.4 (or inverse) with uniform steps. Evaluation results on CIFAR-100 are shown in Fig.~\ref{fig:ablation} (left). Several observations can be made:
% 1) The vanilla sample selection strategy results in a drastic drop of accuracy, suggesting the necessity of our sample \emph{weighting} mechanism;
% 2) The gain of hand-crafted weight values is limited, indicating the advantage of our \emph{learning}-based weighting approach;
% 3) Interestingly, our joint optimization outperforms weighting with a frozen WPN. This suggests that it is essential to \emph{dynamically} adjust the weights imposed on the loss at different training stages.

\begin{itemize}
\item The vanilla sample selection strategy results in a drastic drop in accuracy, suggesting the necessity of our sample \emph{weighting} mechanism;
\item The gain of hand-crafted weight values is limited, indicating the advantage of our \emph{learning}-based weighting approach;
\item Interestingly, our joint optimization outperforms weighting with a frozen WPN. This suggests that it is essential to \emph{dynamically} adjust the weights imposed on the loss at different training stages.
\end{itemize}

\noindent\textbf{Meta-learning objective.} Our designed meta loss in Eq.~(\ref{meta_loss}) encourages every exit to correctly recognize the data \emph{subset} that is most likely handled by the exit in dynamic inference. To clarify the effectiveness of this meta objective, we keep the learning procedure unchanged and substitute Eq.~(\ref{meta_loss}) with the classification loss on the whole meta set for each exit. The results are shown in Fig.~\ref{fig:ablation} (middle). While this variant (line Vanilla Objective) outperforms the baseline when the computation is relatively low, our objective achieves higher accuracy when larger computational budget is available. This suggests that emphasizing \emph{hard} samples for deep classifiers is crucial to improving their performance.
% 1) the meta-learning procedure itself with the vanilla objective can barely improve the network performance; 2) the proposed meta loss is significantly superior to the vanilla objective, which ignores the adaptive inference behavior. 

% \noindent\textbf{The normalization approach.} As mentioned in Sec. \ref{sec_meta_objective}, three normalization formats could be performed on the weight perturbation $\mathbf{\tilde{w}}$: \emph{exit} ($\sum_{k=1}^K \tilde{w}_i^{(k)}=0, \forall i$), \emph{batch} ($\sum_{i=1}^B \tilde{w}_i^{(k)}=0, \forall k$), and \emph{matrix} ($\sum_{i=1}^B\sum_{k=1}^K \tilde{w}_i^{(k)}=0$). We empirically compared these normalization ways, and the results are presented in Fig.~\ref{fig_ablation_metaAll_col_row_mat} (b). It can be observed that the adopted normalization along the \emph{exit} dimension is superior to the other two variants. The visualization results in Fig.~\ref{fig_vis} demonstrate that the learned weights of different exits are clustered in different locations. The weight visualization results for the other two normalization approaches are provided in the supplementary material.

\noindent\textbf{The input of WPN.} In addition to the classification loss, the confidence value can also be leveraged to produce the weight perturbations due to its role in making early exiting decisions. We test three types of input: 1) loss only; 2) confidence only; 3) the concatenation of loss and confidence. The accuracy-computation curves are shown in Fig.~\ref{fig:ablation} (right). It can be found that the adoption of loss is essential for our method, and the inclusion of confidence could be harmful. We hypothesize that by reflecting the information of \textit{both} network prediction and ground truth, the loss serves as a better candidate for WPN input. The ablation study of the WPN design is presented in the supplementary material. 
% Note the confidence value only shows information of network prediction, making it less informative than the loss value.
% The comparison results in the dynamic inference setting are provided in the supplementary material.

\begin{figure}[t]
  \centering
  \begin{subfigure}[b]{0.475\textwidth}
      \centering
      \includegraphics[width=\textwidth]{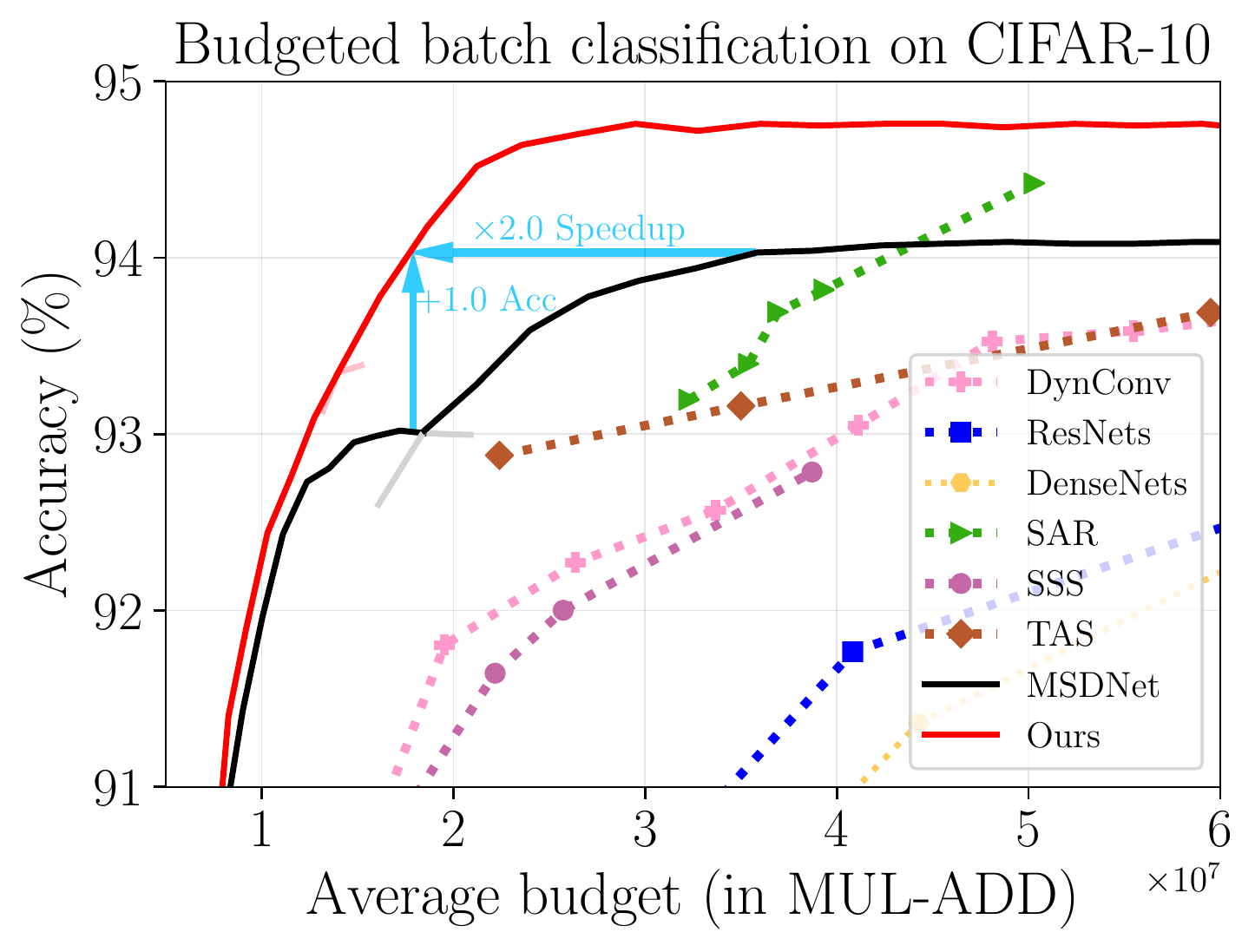}
  \end{subfigure}
  % \hfill
  \begin{subfigure}[b]{0.475\textwidth}
      \centering
      \includegraphics[width=\textwidth]{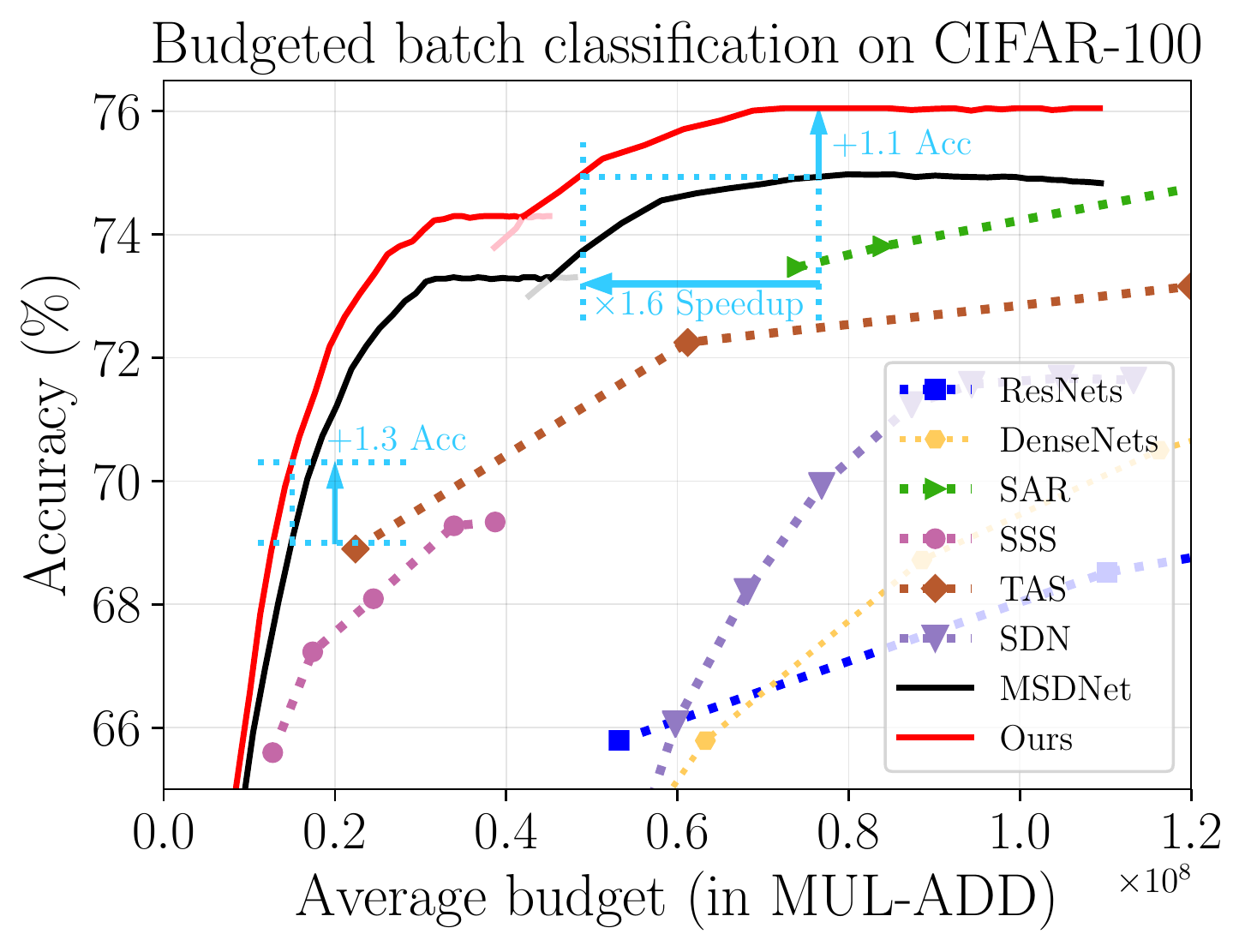}
  \end{subfigure}
%   \vskip -0.1in
  \caption{\textbf{Dynamic inference results} on CIFAR-10 (left) and CIFAR-100 (right)}
    \label{msd_cifar_main_results}
%   \vskip -0.25in
\end{figure}

\subsection{CIFAR results}\label{sec_cifar_results}
In this subsection, we report the results on CIFAR in both dynamic inference and anytime prediction settings following previous works \cite{huang2017multi,li2019improved,yang_resolution_2020}.
We first apply our approach to MSDNet \cite{huang2017multi}, and then compare it with the training techniques in \cite{li2019improved}. The method is further validated on the RANet architecture \cite{yang_resolution_2020}.

\noindent\textbf{Dynamic inference results.} We apply our training strategy on MSDNet with 5 and 7 exits and compare with three groups of competitive baseline methods: classic networks (ResNet\cite{he2016deep}, DenseNet \cite{huang2017densely}), pruning-based approaches (Sparse Structure Selection (SSS) \cite{huang2018data}, Transformable Architecture Search (TAS) \cite{dong2019network}), and dynamic networks (Shallow-Deep Networks (SDN) \cite{kaya_shallow-deep_2019}, Dynamic Convolutions (DynConv) \cite{verelst_dynamic_2020}, and Spatially Adaptive Feature Refinement (SAR) \cite{SAR_TIP}).

\begin{table}[t]
  % \vskip -0.05in
    \begin{center}
    \caption{\textbf{Anytime prediction results} of a 7-exit MSDNet on CIFAR-100}\label{anytime_imta_ours}
    \resizebox{\linewidth}{!}{
        \begin{tabular}{c c c c c c c c c c}
          \toprule
      \multicolumn{2}{c}{Exit index} & 1 & 2 & 3 & 4 & 5 & 6 & 7 \\
      \midrule
      \multicolumn{2}{c}{Params ($\times 10^6$)} & 0.30 & 0.65 & 1.11 & 1.73 & 2.38 & 3.05 & 4.00 \\
      \multicolumn{2}{c}{Mul-Add ($\times 10^6$)} & 6.86 & 14.35 & 27.29 & 48.45 & 76.43 & 108.90 & 137.30 \\
      \midrule
      \multirow{4}*{Accuracy} & MSDNet \cite{huang2017multi} & 61.07 & 64.55 & 67.00 & 69.97 & 72.55 & 74.01 & 74.50 \\
       &\cellcolor{lightgray!50} Ours & \cellcolor{lightgray!50}\textbf{62.47} \scriptsize{($\uparrow$1.30)} &\cellcolor{lightgray!50} \textbf{66.32} \scriptsize{($\uparrow$1.77)} &\cellcolor{lightgray!50} \textbf{68.10} \scriptsize{($\uparrow$1.10)} &\cellcolor{lightgray!50} \textbf{71.29} \scriptsize{($\uparrow$1.32)} &\cellcolor{lightgray!50} \textbf{73.21} \scriptsize{($\uparrow$0.66)} &\cellcolor{lightgray!50} \textbf{74.87} \scriptsize{($\uparrow$0.86)} &\cellcolor{lightgray!50} \textbf{75.81} \scriptsize{($\uparrow$1.31)} \\
      \cmidrule{2-9}
      & IMTA \cite{li2019improved} & 60.29 & 64.86  & 69.09 & 72.71 & 74.47 & 75.60 & 75.19 \\
      & \cellcolor{lightgray!50} Ours + IMTA \cite{li2019improved} &\cellcolor{lightgray!50} \textbf{62.26} \scriptsize{($\uparrow$1.97)} &\cellcolor{lightgray!50} \textbf{67.18} \scriptsize{($\uparrow$2.32)} &\cellcolor{lightgray!50} \textbf{70.53} \scriptsize{($\uparrow$1.44)} &\cellcolor{lightgray!50} \textbf{73.10} \scriptsize{($\uparrow$0.39)} &\cellcolor{lightgray!50} \textbf{74.80} \scriptsize{($\uparrow$0.33)} &\cellcolor{lightgray!50} \textbf{76.05} \scriptsize{($\uparrow$0.45)} &\cellcolor{lightgray!50} \textbf{76.31} \scriptsize{($\uparrow$1.12)} \\
      \bottomrule
      \end{tabular}
      }
      \end{center}
    %   \vskip -0.3in
\end{table}

\begin{figure}[t]
  \centering
  \begin{subfigure}[b]{0.475\textwidth}
      \centering
      \includegraphics[width=\textwidth]{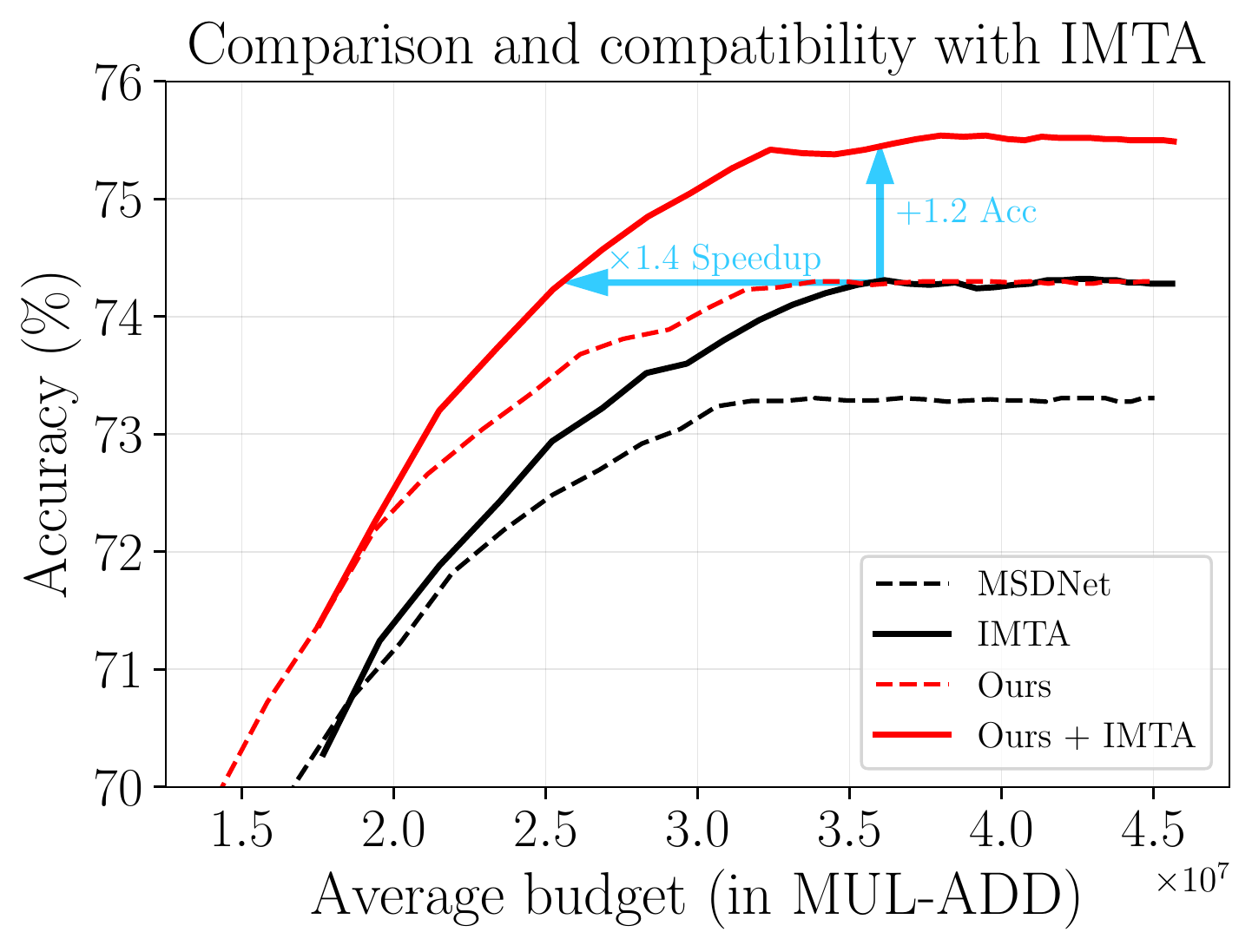}
  \end{subfigure}
  % \hfill
  \begin{subfigure}[b]{0.475\textwidth}
      \centering
      \includegraphics[width=\textwidth]{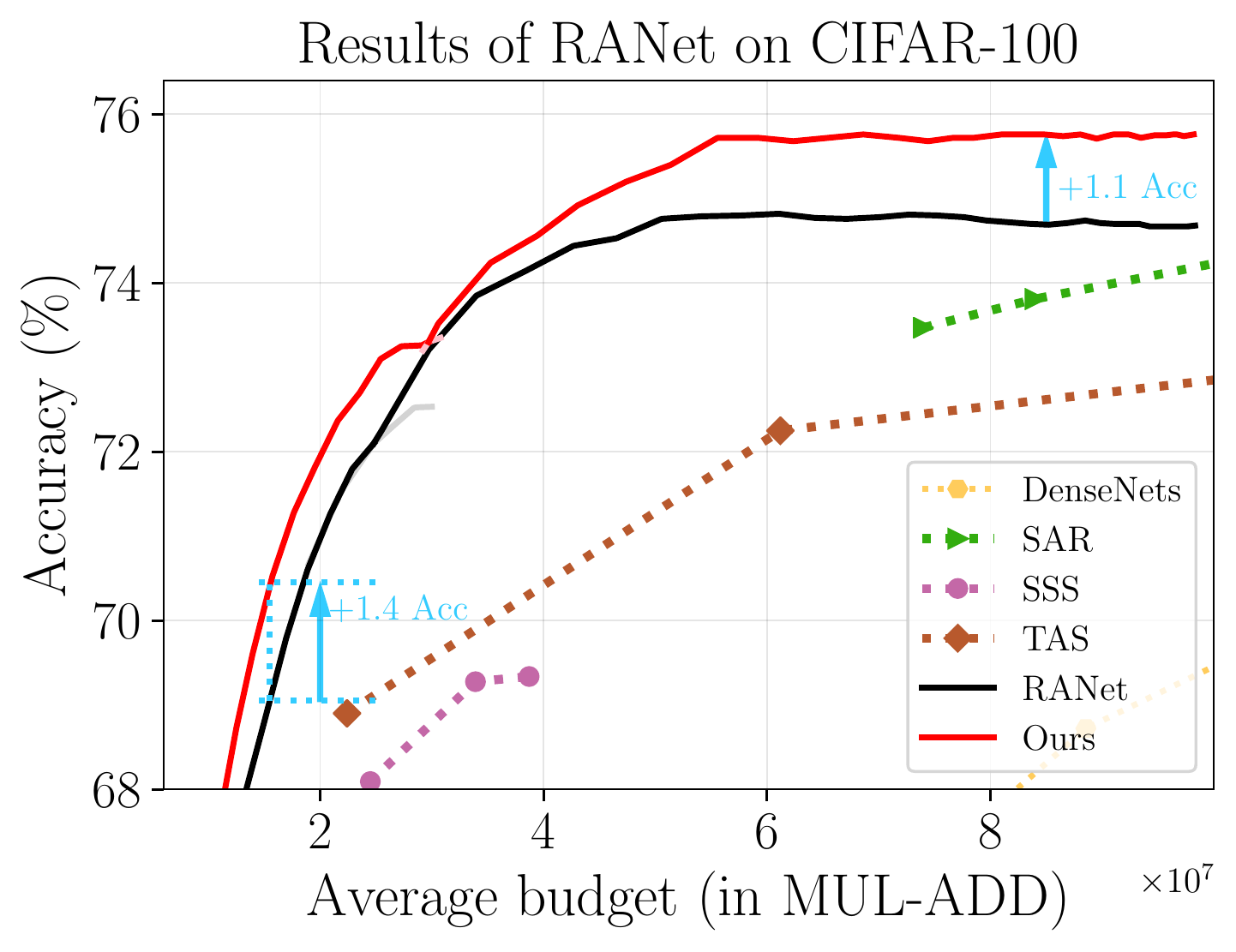}
  \end{subfigure}
%   \vskip -0.1in
  \caption{\textbf{Dynamic inference results on CIFAR-100.} Left: comparison with IMTA \cite{li2019improved}. Right: results with two different sized RANets.}
%   suggests that our weighting method outperforms IMTA when the computational budget is relatively low. It can further improve the accuracy when applied together with IMTA. 
     \label{fig_imta_ra}
%   \vskip -0.15in
\end{figure}

\begin{table}
% \vskip -0.1in
  \begin{center}
  \caption{\textbf{Anytime prediction results} of a 6-exit RANet on CIFAR-100}\label{ra_anytime}
  \resizebox{\linewidth}{!}{
      \begin{tabular}{c c c c c c c c c}
        \toprule
    \multicolumn{2}{c}{Exit index} & 1 & 2 & 3 & 4 & 5 & 6 \\
    \midrule
    \multicolumn{2}{c}{Params ($\times 10^6$)} & 0.36 & 0.90 & 1.30 & 1.80 & 2.19 & 2.62 \\
    \multicolumn{2}{c}{Mul-Add ($\times 10^6$)} & 8.37 & 21.79 & 32.88 & 41.57 & 53.28 & 58.99 \\
    \midrule
    \multirow{2}*{Accuracy} & RANet \cite{yang_resolution_2020}  & 63.41 & 67.36 & 69.62 & 70.21 & 71.00 & 71.78 \\
       & \cellcolor{lightgray!50}Ours      & \cellcolor{lightgray!50}\textbf{65.33} \scriptsize{($\uparrow$1.92)} & \cellcolor{lightgray!50}\textbf{68.69} \scriptsize{($\uparrow$1.30)} &\cellcolor{lightgray!50} \textbf{70.36} \scriptsize{($\uparrow$0.74)} &\cellcolor{lightgray!50}\textbf{70.80} \scriptsize{($\uparrow$0.59)} &\cellcolor{lightgray!50}\textbf{72.57} \scriptsize{($\uparrow$1.57)} &\cellcolor{lightgray!50}\textbf{72.45} \scriptsize{($\uparrow$0.67)} \\
    \bottomrule
    \end{tabular}
    }
    \end{center}
    % \vskip -0.4in
\end{table}

\begin{figure}[h]
  \centering
  \begin{subfigure}[b]{0.495\textwidth}
    \centering
    \includegraphics[width=\textwidth]{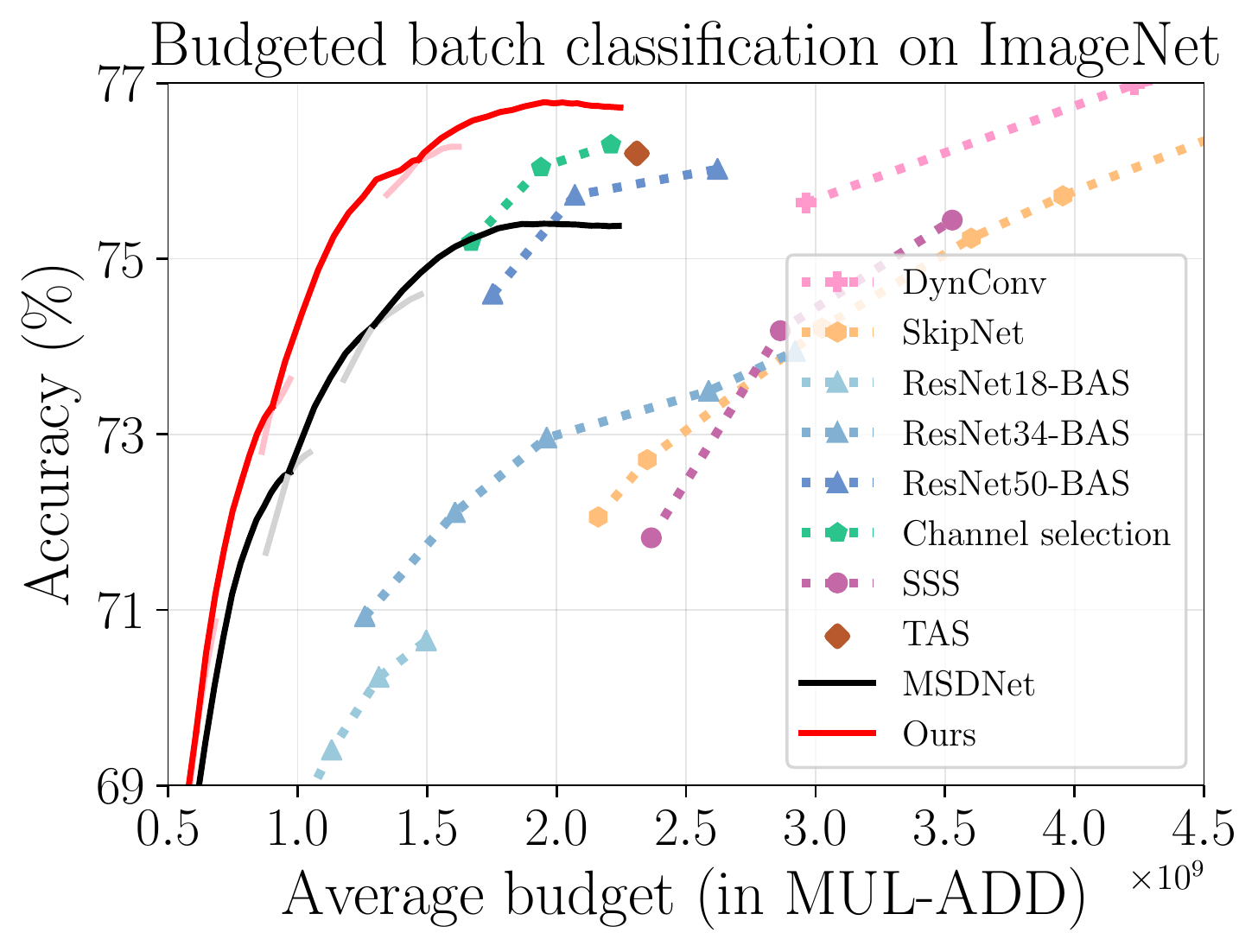}
  \end{subfigure}
  \begin{subfigure}[b]{0.475\textwidth}
      \centering
      \includegraphics[width=\textwidth]{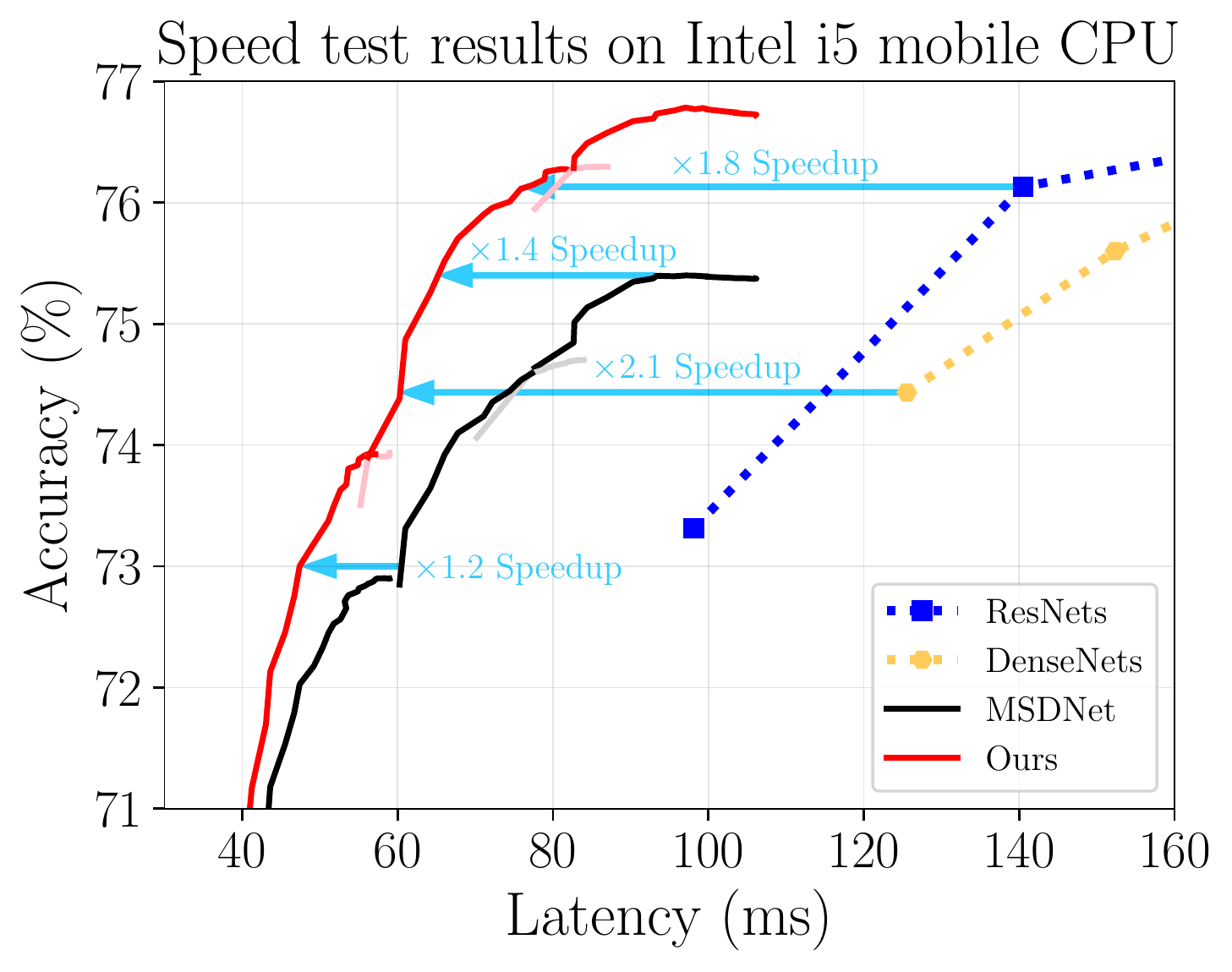}
  \end{subfigure}
  \hfill
  \begin{subfigure}[b]{0.475\textwidth}
      \centering
      \includegraphics[width=\textwidth]{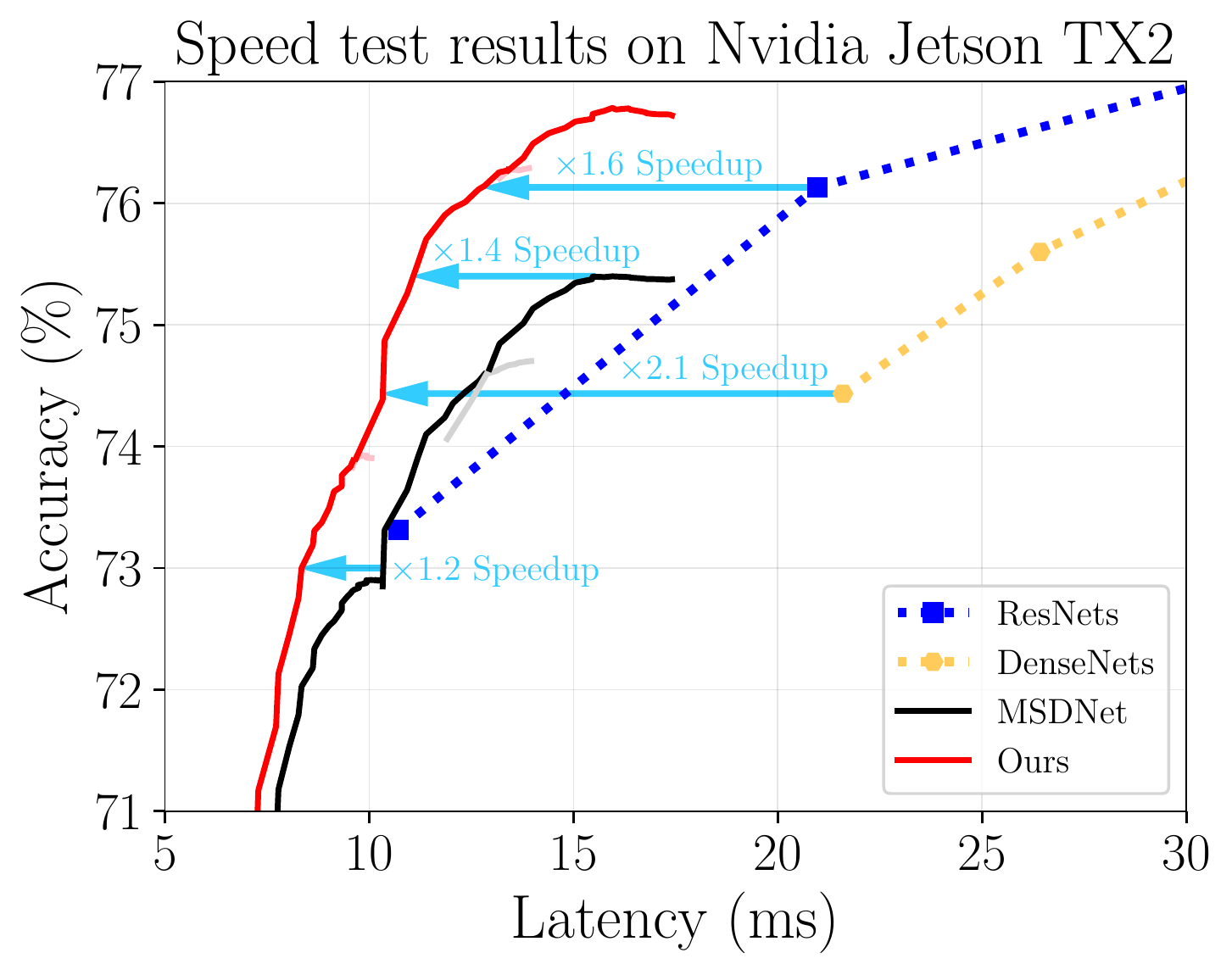}
  \end{subfigure}
  \begin{subfigure}[b]{0.475\textwidth}
      \centering
      \includegraphics[width=\textwidth]{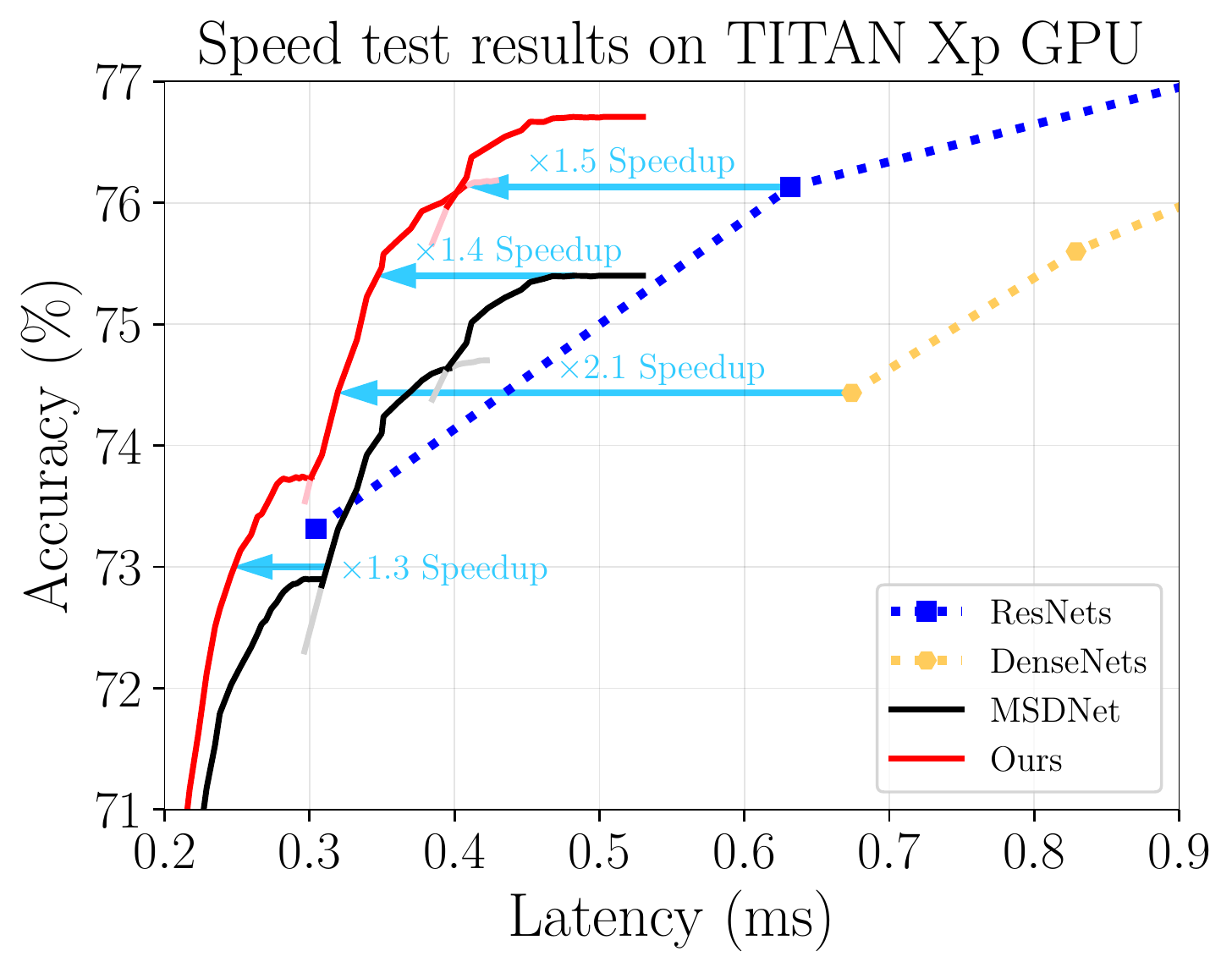}
  \end{subfigure}
%   \vskip -0.15in
  \caption{\textbf{ImageNet results.} Top left: accuracy-computation curves. Others: accuracy-latency curves on different hardware platforms.}
%   Top right: accuracy-latency curves on Intel i5 mobile CPU. Bottom left: accuracy-latency curves on TX2. Bottom right: accuracy-latency curves on TITAN Xp GPU.}
     \label{IN_results}
%   \vskip -0.2in
\end{figure}
\begin{figure}[h]
  \centering
%   \vskip -0.1in
  \begin{subfigure}[b]{0.46\textwidth}
    \centering
    \includegraphics[width=\textwidth]{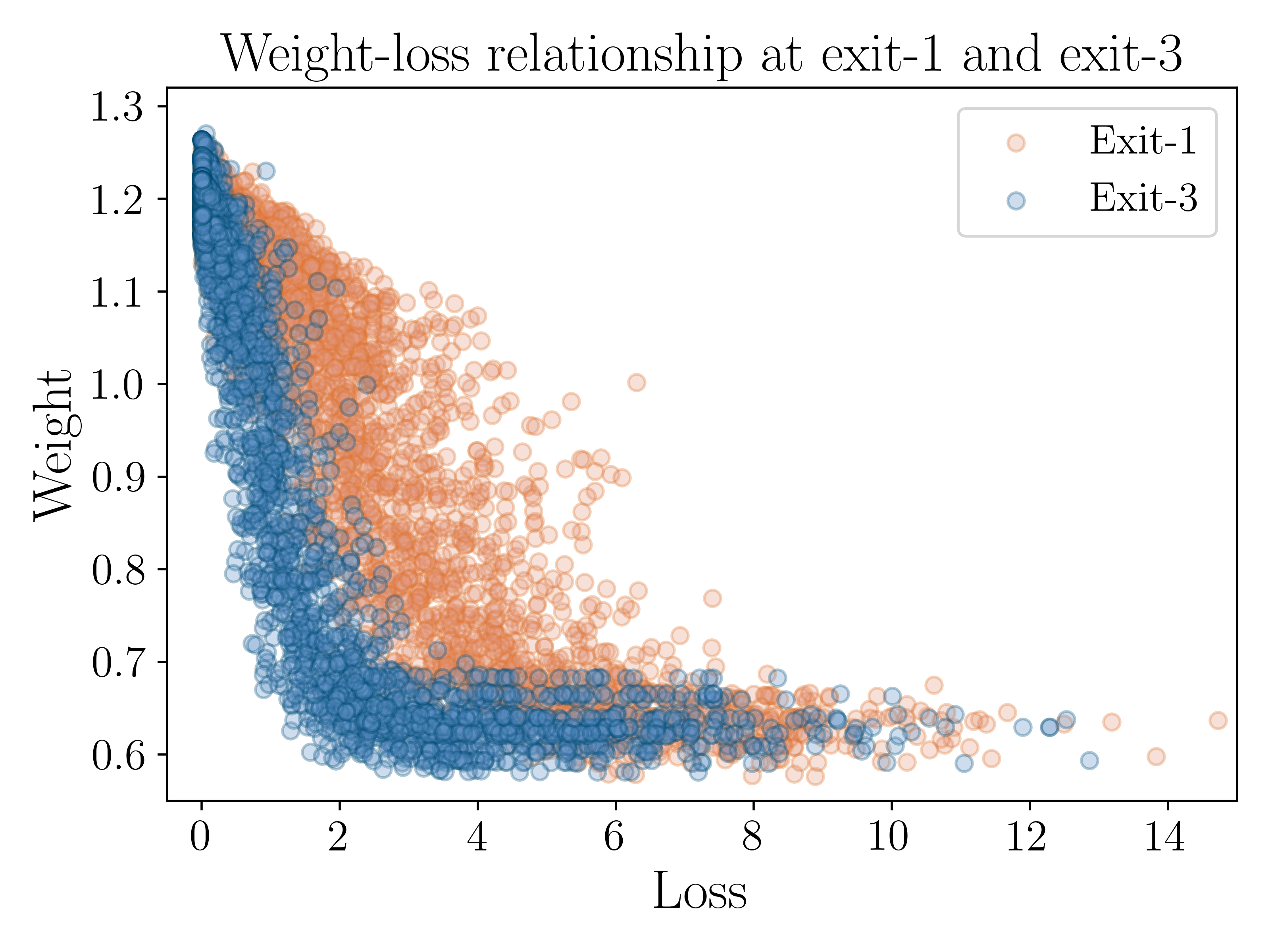}
  \end{subfigure}
  \begin{subfigure}[b]{0.4425\textwidth}
      \centering
      \includegraphics[width=\textwidth]{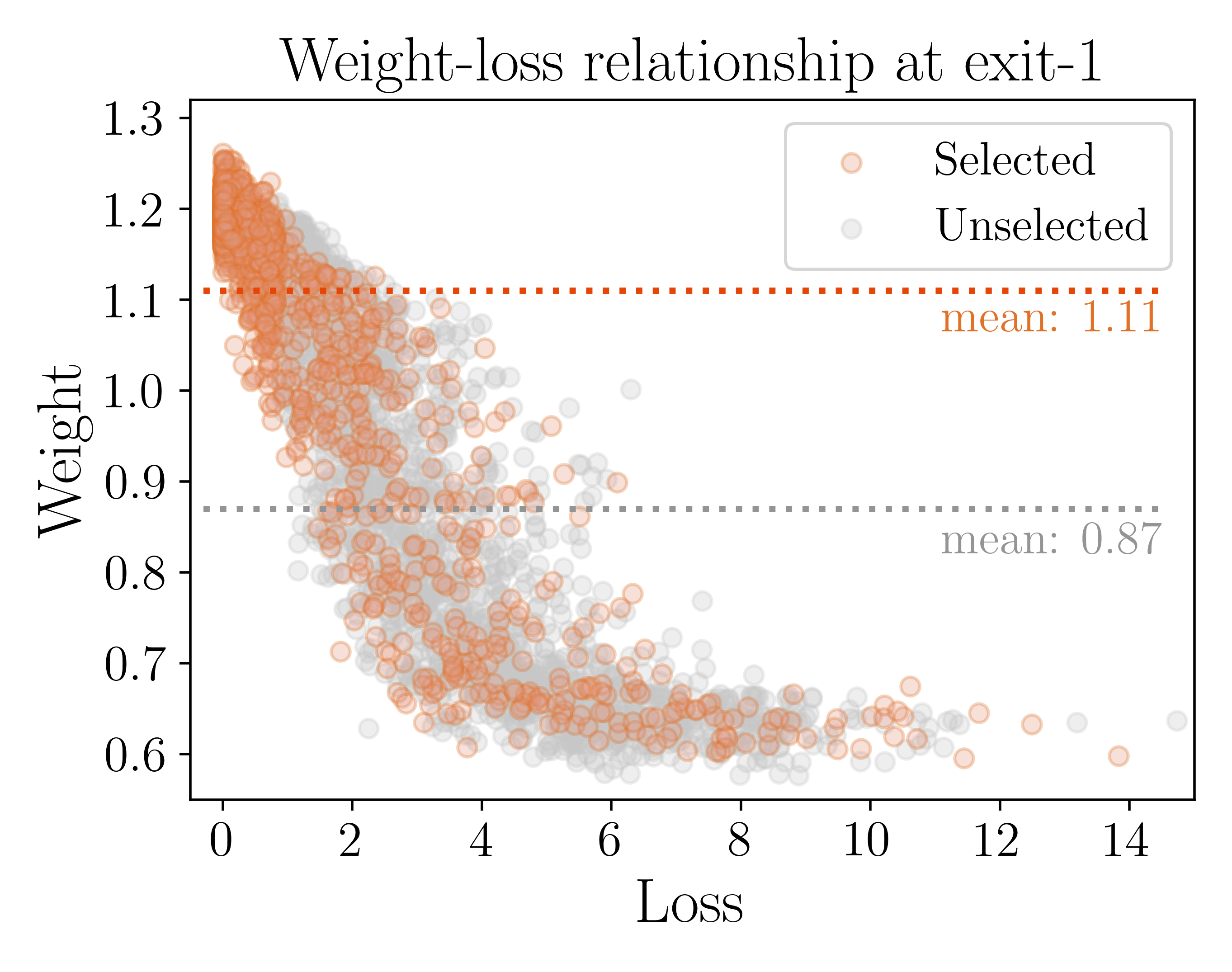}
  \end{subfigure}
%   \vskip -0.2in
  \caption{\textbf{Visualization results} of the weight-loss relationship on ImageNet}
     \label{fig_vis}
%   \vskip -0.1in
\end{figure}

In the dynamic inference scenario, we present the accuracy-computation (measured by Mul-Adds) curve in Fig.~\ref{msd_cifar_main_results} (left: CIFAR-10, right: CIFAR-100). From the results, we can observe that the proposed weighting method consistently improves the performance of MSDNet at various computational budgets. When applied to the 5-exit model, our weighting mechanism obtains significant boosts with the CIFAR-100 Top-1 accuracy increased by about 1.3\% when evoking around 15M Mul-Adds. For the 7-exit MSDNet, our model only uses half of the original budget to achieve $\sim$94.0\% Top-1 accuracy on CIFAR-10. 
% When the resource budget allows most samples to be output at deep exits, the overall accuracy of the network can be improved by about 1.3\%.

\noindent\textbf{Anytime prediction results.}\label{sec_anytime_results} 
We also report the accuracy of each exit on the whole test set in the anytime prediction setting. From the results in Tab.~\ref{anytime_imta_ours} (a 7-exit MSDNet \cite{huang2017multi}) and Tab.~\ref{ra_anytime} (a 6-exit RANet \cite{yang_resolution_2020}), we surprisingly find that although our meta objective encourages each exit to focus on only a \emph{subset} of samples, the performance on the \emph{whole} test set is consistently increased by a large margin. 
This phenomenon could bring some insights into the optimization of deep networks:
% Despite the conventional wisdom focuses on those \emph{hard} samples which are close to the decision boundary, recent work \cite{zhao2021well} has shown that the well-classified samples may be underestimated in training deep models. 
1) for the shallow exits, emphasizing a small subset of \emph{easy} samples with \emph{high confidence} benefits their generalization performance on the whole dataset \cite{wang2021survey_cl}; 2) for the deeper exits, our objective forces them to focus on the \emph{hard} samples, which cannot be confidently predicted by previous classifiers. Such a ``challenging'' goal could further improve their capability of approximating more complex classification boundaries. This coincides with the observation in our ablation studies (Fig.~\ref{fig:ablation} middle): encouraging each exit to correctly classify a \emph{well-selected} subset of training samples is preferable.

\noindent\textbf{Comparison and compatibility with IMTA.}\label{sec_comp_imta}
Improved techniques for training adaptive networks (IMTA) \cite{li2019improved} are proposed to stabilize training and facilitate the collaboration among exits. These techniques are developed for the \emph{optimization procedure} and ignore the adaptive inference behavior. 
In contrast, our meta-learning based method improves the \emph{objective optimization} and takes the inference behavior into account. 
We empirically compare our approach to IMTA \cite{li2019improved}, and further combine them to evaluate the compatibility. 
% We follow the pretrain-finetuning process in IMTA \cite{li2019improved}, and our meta-learning procedure is included in both phases. 

The results are presented in Fig.~\ref{fig_imta_ra} (right, for dynamic inference) and Tab.~\ref{anytime_imta_ours} (row 6 \& 7, for anytime prediction). These results validate that 1) our weighting strategy achieves higher performance, especially at low computational budges; 2) combining our method with IMTA further improves the performance. In particular, note that the accuracy of deep exits can be boosted by a large margin.

% \begin{table}
%   \begin{center}
%   \resizebox{\linewidth}{!}{
%       \begin{tabular}{c c c c c c c c c c}
%         \toprule
%     % \multicolumn{2}{c}{Exit index} & 1 & 2 & 3 & 4 & 5 & 6 \\
%     % \midrule
%     % \multicolumn{2}{c}{Params ($\times 10^6$)} & 0.36 & 0.90 & 1.30 & 1.80 & 2.19 & 2.62 \\
%     \multicolumn{2}{c}{Mul-Add ($\!\times\!10^9$)} & 0.35 & 0.41 & 0.48 & 0.54 & 0.64 & 0.75 & 1.11\\
%     \midrule
%      \ Top-1 & Baseline  & 58.66 & 61.45 & 64.40 & 66.53 & 69.47 & \textbf{71.36} & \textbf{73.35} \\
%      \ Acc   & \cellcolor{lightgray!50}Ours & \cellcolor{lightgray!50}\textbf{60.56} &\cellcolor{lightgray!50}\textbf{63.46}&\cellcolor{lightgray!50}\textbf{65.73} &\cellcolor{lightgray!50}\textbf{67.49} &\cellcolor{lightgray!50}\textbf{69.66} &\cellcolor{lightgray!50}71.21 & \cellcolor{lightgray!50}73.28\\
%     \bottomrule
%     \end{tabular}}
%     \end{center}
%     \vskip -0.25in
%     \caption{Budgeted batch classification results on ImageNet.}
%     \label{tab_imgnet}
%     \vskip -0.15in
% \end{table}

% We first compare our approach with IMTA. Note that IMTA requires finetuning a pretrained model, which increase the optimization iterations. For fair comparison, we extend our epoch number to be equal to that in \cite{li2019improved}. The results in Tab.~\ref{anytime_imta_ours} (row 7) show that with the same amount of training iterations, our method is superior to IMTA.

\noindent\textbf{Results on RANet.} To evaluate the generality of our method, we conduct experiments on another representative multi-exit structure, RANet \cite{yang_resolution_2020}. The anytime prediction performance of a 6-exit RANet is shown in Tab.~\ref{ra_anytime}, and the dynamic inference results of two different-sized RANets are illustrated in Fig.~\ref{fig_imta_ra} (right). These results suggest that the proposed approach consistently improves the trade-off between accuracy and computational cost of RANet. Notably, when the Mul-Add is around 15M, our model improves the Top-1 accuracy significantly ($\sim\!1.4\%$). The strong performance on MSDNet and RANet indicates that our weighting algorithm is sufficiently \emph{general} and \emph{effective}.

\subsection{ImageNet results}\label{sec_imagenet}
Next, we evaluate our method on the large-scale ImageNet dataset \cite{deng2009imagenet}. We test three different sized MSDNets following \cite{huang2017multi} and compare them with competitive baselines, including pruning-based approaches (the aforementioned SSS \cite{huang2018data} and TAS \cite{dong2019network}, and Filter Pruning via Geometric Median (Geom) \cite{he2019filter}) and dynamic networks (SkipNet \cite{wang2018skipnet}, Batch-Shaping (BAS) \cite{bejnordi2019batch}, Channel Selection \cite{herrmann2018end} and DynConv \cite{verelst_dynamic_2020}). We report the results in the dynamic inference setting here. Anytime prediction results can be found in the supplementary material.

\noindent\textbf{Accuracy-computation result.} We present the accurracy-computation results on ImageNet in Fig.~\ref{IN_results} (top left). The plot shows that the weighting mechanism consistently improves the accuracy-computation trade-off of MSDNets. Even though some competing models surpass the baseline MSDNet trained without sample weighting, our method successfully outperforms these competitors.

\noindent\textbf{Accuracy-latency result.} We benchmark the efficiency of our models on three types of computing devices: Intel i5-1038NG7 mobile CPU, Nvidia Jetson TX2 (a resource-constrained embedded AI computing device) and TITAN Xp GPU. The testing batch sizes are set as 64 for the first two devices and 512 for the last. The testing images have resolutions of $224\!\times\!224$ pixels. The accuracy-latency curves plotted in Fig.~\ref{IN_results} demonstrate the significant improvement of our method across all computing platforms. For instance, our model only takes 70\% of the original MSDNet's computation time to achieve 75.4\% accuracy ($1.4\times\!$ speed-up).
% , the practical efficiency of MSDNet could be improved by 1.4 times. 
% With the same computational cost, the accuracy of the largest model is increased by 1.4\%.

\noindent\textbf{Visualization.} We visualize the weights predicted by WPN toward the end of training in Fig.~\ref{fig_vis} \footnote{We set $q\! =\!0.5$ in training, and therefore the proportion of output samples at 5 exits follows an exponential distribution of $[0.52,0.26,0.13,0.06,0.03]$.}. The left plot presents the weight-loss relationship at exit-1 and exit-3. The right plot shows the weights of the samples selected/not selected by exit-1 in meta-data allocation (Sec.~\ref{sec:sample-weighting}). Several observations can be made:
1) Since more samples are allocated to early exits, the weights at exit-1 are generally larger than those at exit-3. This result suggests that the proposed method successfully learns the relationship between exits;
2) Both exits tend to emphasize the samples with a smaller loss. This coincides with the observation in \cite{zhao2021well} which explains the importance of the well-classified (\emph{easy}) samples in the optimization of deep models;
3) From weights at exit-1, we can observe that the samples with high prediction confidence (which would be selected by exit-1 in meta-data allocation) generally have weights larger than samples with lower confidence (which would be allocated to deeper classifiers). This suggests that the proposed method learns to recognize the samples' exit at different classifiers.

% \begin{itemize}
%     \item Since more samples are allocated to early exits, the weights at exit-1 are generally larger than those at exit-3. This result suggests that the proposed method successfully learns the relationship between exits;
    
%     \item Both exits tend to emphasize the samples with smaller loss. This coincides with the observation in \cite{zhao2021well} which explains the importance of the well-classified (\emph{easy}) samples in the optimization of deep models;
    
%     \item From weights at exit-1, we can observe that the samples with high prediction confidence (which would be selected by exit-1 in meta-data allocation) generally have weights larger than samples with lower confidence (which would be allocated to deeper classifiers). This suggests that the proposed method learns to recognize the samples exit at different classifiers.
    
%     % the weights have two clusters (Fig.~\ref{fig_vis} (b)). Especially, a number of correctly recognized samples (with loss nearing 0) are imposed with small weights. The other cluster still has considerable loss together with large weights. Note that in the meta objective (Eq.~(\ref{meta_loss})), we only emphasize the samples which are predicted with high confidence. However, our weighting mechanism does not yield large weights for all the confident samples. Instead, it \emph{learns} to identify and correct those \emph{over-confident} predictions, to reduce the risk of these incorrect predictions being output with high confidence.
% \end{itemize}

\begin{figure}[t]
% \vskip -0.05in
  \centering
    \includegraphics[width=\linewidth]{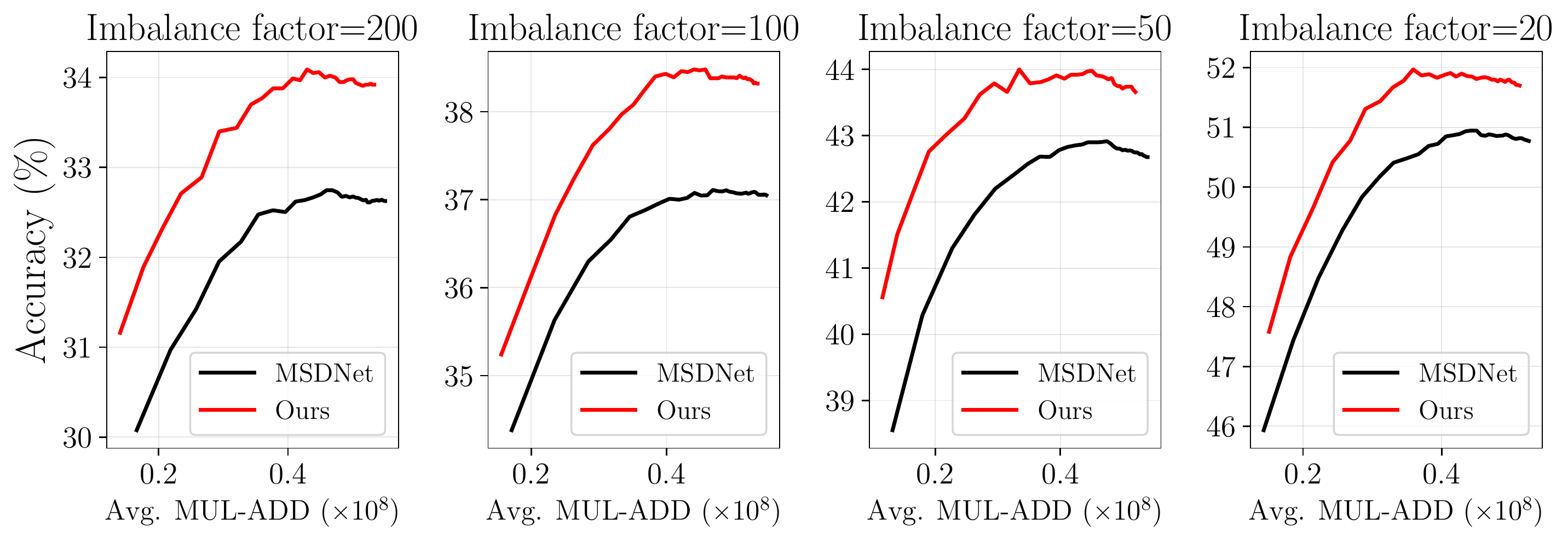}
    % \vskip -0.1in
    \caption{\textbf{Results on long-tailed CIFAR-100} with different imbalance factors}
    \label{fig_imbalance}
    % \vskip -0.2in
\end{figure}

\subsection{Class imbalance results}\label{sec_imbalance}
The proposed method is finally evaluated in the class imbalance setting. On the long-tailed CIFAR-100 \cite{cui2019class} dataset, we test with four imbalance factors (200, 100, 50, and 20), where the imbalance factor is defined as the number of training samples in the largest class divided by the smallest. 
% Our meta objective is still defined as in Eq.~(\ref{meta_loss}) and calculated on the meta-data. 
The budgeted batch classification performance illustrated in Fig.~\ref{fig_imbalance} shows that our weighting method consistently outperforms the conventional training scheme by a large margin. 
% Especially, when the imbalance factor is 200, the accuracy of MSDNet could be improved by a large margin ($>1\%$) with varying computational budgets.
\section{Conclusion}
\label{sec:conclusion}
In this paper, we propose a meta-learning based sample weighting mechanism for training dynamic early-exiting networks. Our approach aims to bring the test-time adaptive behavior into the training phase by sample weighting. We weight the losses of different samples at each exit by a weight prediction network. This network is jointly optimized with the backbone network guided by a novel meta-learning objective. The proposed weighting scheme can consistently boost the performance of multi-exit models in both anytime prediction and budgeted classification settings. 
% Moreover, we show that the weighting strategy is compatible with different network architectures.
Experiment results validate the effectiveness of our method on the long-tailed image classification task.
% The potential limitations are discussed in the supplementary material.
% Future research may focus on improving the training efficacy of other types of adaptive models, such as spatial-wise and temporal-wise dynamic networks.

\noindent\textbf{Acknowledgement.} This work is supported in part by National Key R\&D Program of China (2020AAA0105200), the National Natural Science Foundation of China under Grants 62022048, Guoqiang Institute of Tsinghua University and Beijing Academy of Artiﬁcial Intelligence. We also appreciate the generous donation of computing resources by High-Flyer AI.

% \clearpage
\bibliographystyle{splncs04}
\bibliography{reference}
\clearpage
\appendix
% \title{Appendix} % R
% \author{}
% \institute{}
% \maketitle
\section*{Appendix}

The organization of the supplementary material is as follows. We first introduce the detailed experiment setting in Sec. \ref{sec:exp_setup}, including the datasets, model configuration, hyper-parameter selection and training details. Next, we provide the anytime prediction results of ImageNet in Sec. \ref{sec:anytime}. Finally, the ablation studies of the \emph{weight prediction network} (WPN) design are presented in Sec. \ref{sec:WPN}.

\section{Experimental Setup}
\label{sec:exp_setup}

\subsection{Datasets}

CIFAR-10 and CIFAR-100~\cite{Krizhevsky09learningmultiple}  contain 10 and 100 classes of natural scene objects, respectively. Both datasets contain 50,000 training images and 10,000 test images. Each image has a resolution of $32\times32$ pixels. 
ImageNet~\cite{deng2009imagenet} contains 1.2 million training images and 50,000 validation images in 1000 classes. 
Following \cite{huang2017densely}, we use the basic data augmentation for CIFAR-10, CIFAR-100 and ImageNet.
The long-tailed CIFAR dataset~\cite{cui2019class}, unlike CIFAR and ImageNet, does not have a well-balance class distribution. It reduces the number of training samples in each class based on an exponential function
$N'_c=N_c\mu^c, \mu\in(0,1)$, where $c$ is the class index, and $N_c$ is the original training sample number.

\subsection{Model Configuration}

\subsubsection{MSDNet.} We follow the model configurations in \cite{huang2017multi}. For the CIFAR dataset, we use MSDNet with three different scales ($32 \times 32, 16 \times 16, 8 \times 8$). The trained MSDNets have \{5, 7\} classifiers, where their depths are \{15, 28\}, respectively. The $k^{th}$ classifier is attached to the $(\sum_{i=1}^{k} i)^{th}$ layer. On ImageNet, the MSDNet has four scales, with the $
k^{th}$ classifier attaching on the ${(s \times k + 3)}^{th}$ layer $(k=1,...,5)$. The hyper-parameter $s$ controls the total depth and the location of each classifier of the network, and is set as $4,6$ and $7$ to get models with different sizes.

\subsubsection{RANet.} We use the same model structure proposed in \cite{yang_resolution_2020}. In the main paper, we conduct experiments on two RANets. They both have three scales but have three and four base features, respectively. Detailed configuration is as follows:

\textbf{Model-C-1}: The sizes of three base features are $32 \times 32, 16 \times 16, 8 \times 8$. Three sub-networks respectively  corresponding to the base features have 6, 4, 2 convolutional blocks. We use the linear growth step mode to control the number of layers in each block, which means the number of layers in a block is added 2 to the previous one, and the base number of layers is 2. The channel numbers in these base features are 16, 32, 64, which are input channels numbers for different sub-networks. The growth rates of the three sub-networks are 6, 12, 24. The Model-C-1 has 6 classifiers in total.

\textbf{Model-C-2}: The sizes of four base features are $32 \times 32, 16 \times 16, 16 \times 16, 8 \times 8$. These four sub-networks respectively  corresponding to these base features have 8, 6, 4, 2 convolutional blocks. Linear growth step mode is used to control the depth of each block. The number of input channels and growth rates are 16, 32, 32, 64 and 6, 12, 12, 24, respectively. The Model-C-2 has 8 classifiers in total.

\subsection{Hyper-parameter selection}
For the budget controlling variable $q$, we hold out an extra validation set from the training data and tune $q$ on it. We find 0.75 work well for CIFAR, and 0.5, 0.75 work well for ImageNet. The magnitude of perturbation $\delta$ (Sec. 3.2,
weighted classification loss) is set as 0.8 on CIFAR and 0.3 on ImageNet.

\subsection{Training details}

We use the stochastic gradient descent (SGD) to train the backbone parameters $\mathbf{\Theta}_f$. The batch sizes are set to 1024 and 2048 for CIFAR and ImageNet, respectively. The learning rate for backbone parameters starts with $0.1\times$batch size/64 for CIFAR and $0.1\times$batch size/256 for ImageNet, decaying with a cosine shape. We use a momentum of 0.9 and a weight decay of $1\times 10^{-4}$ for both datasets. All the models (with or without our weighting mechanism) are trained for 300 epochs on CIFAR and 100 epochs on ImageNet. 

Following \cite{meta_weight_net}, we adopt Adam \cite{kingma2014adam} with an initial learning rate of $1\!\times\! 10^{-4}$ to optimize our WPN $\mathbf{\Theta}_g$. The interval $I$ of updating WPN in Algorithm 1 is set to 1 and 100 for CIFAR and ImageNet, respectively.

In each iteration of our meta-learning algorithm, a mini-batch is chunked into two parts: 
one is used to train the backbone parameters $\mathbf{\Theta}_f$, while the other part is used for training the WPN parameters $\mathbf{\Theta}_g$. We then exchange the two parts to keep the total number of training iterations equal to the baseline strategy. For experiments of IMTA compatibility, we follow the pretrain-finetuning process in IMTA \cite{li2019improved}, and our meta-learning procedure is included in both phases.

Experiments on CIFAR-10, CIFAR-100 and long-tailed CIFAR are conducted on Nvidia Geforce RTX 2080 Ti GPUs, and the training for ImageNet models is conducted on Nvidia Tesla A100 GPUs.

\begin{table}
  \begin{center}
  \caption{\textbf{Anytime prediction results} of a MSDNet with $k=4$ on ImageNet}\label{msd_anytime_s4}
  \resizebox{0.8\linewidth}{!}{
      \begin{tabular}{c c c c c c c c c}
        \toprule
    \multicolumn{2}{c}{Exit index} & 1 & 2 & 3 & 4 & 5 \\
    \midrule
    \multicolumn{2}{c}{Params ($\times 10^6$)} & 4.24 & 8.77 & 13.07 & 16.75 & 23.96 \\
    \multicolumn{2}{c}{Mul-Add ($\times 10^9$)} & 0.34 & 0.69 & 1.01 & 1.25 & 1.36 \\
    \midrule
    \multirow{2}*{Accuracy} & MSDNet  & 59.00 & 66.78 & 70.12 & 71.42 & 72.90 \\
       & \cellcolor{lightgray!50}Ours      & \cellcolor{lightgray!50}\textbf{59.54} \scriptsize{($\uparrow$0.54)} & \cellcolor{lightgray!50}\textbf{67.22} \scriptsize{($\uparrow$0.44)} &\cellcolor{lightgray!50} \textbf{71.03} \scriptsize{($\uparrow$0.91)} &\cellcolor{lightgray!50}\textbf{72.33} \scriptsize{($\uparrow$0.91)} &\cellcolor{lightgray!50}\textbf{73.93} \scriptsize{($\uparrow$1.03)} \\
    \bottomrule
    \end{tabular}
    }
    \end{center}
\end{table}

\begin{table}
  \begin{center}
  \caption{\textbf{Anytime prediction results} of a MSDNet with $k=6$ on ImageNet}\label{msd_anytime_s6}
  \resizebox{0.8\linewidth}{!}{
      \begin{tabular}{c c c c c c c c c}
        \toprule
    \multicolumn{2}{c}{Exit index} & 1 & 2 & 3 & 4 & 5 \\
    \midrule
    \multicolumn{2}{c}{Params ($\times 10^6$)} & 4.24 & 10.78 & 17.84 & 24.58 & 38.68 \\
    \multicolumn{2}{c}{Mul-Add ($\times 10^9$)} & 0.34 & 0.92 & 1.52 & 1.99 & 2.19 \\
    \midrule
    \multirow{2}*{Accuracy} & MSDNet  & 58.69 & 68.75 & 72.35 & 73.59 & 74.63 \\
       & \cellcolor{lightgray!50}Ours      & \cellcolor{lightgray!50}\textbf{60.05} \scriptsize{($\uparrow$1.36)} & \cellcolor{lightgray!50}\textbf{69.13} \scriptsize{($\uparrow$0.38)} &\cellcolor{lightgray!50} \textbf{73.33} \scriptsize{($\uparrow$0.98)} &\cellcolor{lightgray!50}\textbf{75.19} \scriptsize{($\uparrow$1.60)} &\cellcolor{lightgray!50}\textbf{76.30} \scriptsize{($\uparrow$1.67)} \\
    \bottomrule
    \end{tabular}
    }
    \end{center}    
\end{table}

\begin{table}
  \begin{center}
  \caption{\textbf{Anytime prediction results} of a MSDNet with $k=7$ on ImageNet}\label{msd_anytime_s7}
  \resizebox{0.8\linewidth}{!}{
      \begin{tabular}{c c c c c c c c c}
        \toprule
    \multicolumn{2}{c}{Exit index} & 1 & 2 & 3 & 4 & 5 \\
    \midrule
    \multicolumn{2}{c}{Params ($\times 10^6$)} & 4.24 & 11.89 & 20.58 & 29.16 & 47.54 \\
    \multicolumn{2}{c}{Mul-Add ($\times 10^9$)} & 0.34 & 1.06 & 1.81 & 2.42 & 2.68 \\
    \midrule
    \multirow{2}*{Accuracy} & MSDNet  & 58.81 & 69.45 & 73.18 & 74.32 & 75.35 \\
       & \cellcolor{lightgray!50}Ours      & \cellcolor{lightgray!50}\textbf{59.24} \scriptsize{($\uparrow$0.43)} & \cellcolor{lightgray!50}\textbf{69.65} \scriptsize{($\uparrow$0.20)} &\cellcolor{lightgray!50} \textbf{73.94} \scriptsize{($\uparrow$0.76)} &\cellcolor{lightgray!50}\textbf{75.66} \scriptsize{($\uparrow$1.34)} &\cellcolor{lightgray!50}\textbf{76.72} \scriptsize{($\uparrow$1.37)} \\
    \bottomrule
    \end{tabular}
    }
    \end{center}
\end{table}

\section{Anytime prediction results of ImageNet}
\label{sec:anytime}

The ImageNet results in dynamic inference setting have been presented in the paper. In this section, the anytime prediction results of different sized MSDNet are shown in Tab.~\ref{msd_anytime_s4}, Tab.~\ref{msd_anytime_s6}, Tab.~\ref{msd_anytime_s7}, respectively.

\begin{table}
  \caption{{Ablation study of the WPN size on CIFAR-100}.}
  \begin{center}
  \resizebox{0.6\linewidth}{!}{
      \begin{tabular}{c c c c c c c }
        \toprule
    \multicolumn{2}{c}{Exit index} & 1 & 2 & 3 & 4 & 5 \\
    \midrule
    \multicolumn{2}{c}{Multi-Add ($\!\times\!10^6$)} & 6.86 & 14.35 & 27.54 & 41.71 & 58.48  \\
    \midrule
    \multirow{6}*{Top-1 Acc}& Baseline & 61.12 & 63.60 & 69.00 & 71.32 & 72.51  \\
    \cmidrule{2-7}
    & $d$=1, $w$=10  & 62.06 & 65.28  & 70.19  & 71.47  & 72.77  \\
    & $d$=1, $w$=100 & 63.10 & 66.21  & 69.36  & 71.66  & 72.98 \\
    % & $d$=1, $w$=300 & 62.54 & 65.21  & 69.53  & 71.32  & 72.64 \\
    & \cellcolor{lightgray!50} $d$=1, $w$=500 & \cellcolor{lightgray!50} 62.61  & 
    \cellcolor{lightgray!50} 65.71  &
    \cellcolor{lightgray!50} 69.14  &  \cellcolor{lightgray!50} 72.12  & \cellcolor{lightgray!50} 73.36  \\
    & $d$=1, $w$=1000 & 63.06 & 66.07  & 68.60   & 72.10   & 73.25  \\
    & $d$=2, $w$=500 & 62.95 & 65.88  & 69.17  & 71.56  & 73.18 \\
    & $d$=4, $w$=500 & 63.13 & 66.09  & 69.41  & 71.96  & 73.06 \\
    \bottomrule
    \end{tabular}
    }
    \end{center}
    \label{wpn_anytime}
\end{table}

\section{More Ablation Studies}
\label{sec:WPN}

\noindent\textbf{WPN design.} For a given $k$-exit model, the input and output dimensions of the WPN is fixed as $k$. We conduct ablation studies of the WPN \emph{depth} (the number of hidden layers, $d$) and \emph{width} (the number of neurons of the hidden layer, $w$) on CIFAR-100. Note that our default setting in the paper is $d$=1 and $w$=500 (for a 5-exit model). The results in Table~\ref{wpn_anytime} demonstrate that the performance is relatively stable as long as the WPN is not too small. The gain mainly comes from the overall \emph{meta-learning} paradigm and our novel meta \emph{objective}.

% \clearpage
\end{document}